\documentclass{article}
\usepackage{a4wide}

\usepackage{graphicx}
\usepackage{mathptmx}      % use Times fonts if available on your TeX system

\graphicspath{{./}}

\usepackage{makeidx}
\usepackage[utf8]{inputenc}
\usepackage[T1]{fontenc}
\usepackage{dashundergaps}
\usepackage[english]{babel}
\usepackage{subfig}
\usepackage{color}
\usepackage{algorithm}
\usepackage{algorithmicx}
\usepackage{algpseudocode}
\usepackage{booktabs}
\usepackage{amsmath}
\usepackage{amssymb}
\usepackage{url}
\usepackage{subfloat}
\usepackage{tikz}
\usepackage{natbib}

\renewcommand\cite{\citep}

\makeatletter
\newcommand{\algmargin}{\the\ALG@thistlm}
\makeatother
\newlength{\whilewidth}
\algdef{SE}[parWHILE]{parWhile}{EndparWhile}[1]
  {\parbox[t]{\dimexpr\linewidth-\algmargin}{%
     \hangindent\whilewidth\strut\algorithmicwhile\ #1\ \algorithmicdo\strut}}{\algorithmicend\ \algorithmicwhile}%
\algnewcommand{\parState}[1]{\State%
  \parbox[t]{\dimexpr\linewidth-\algmargin}{\strut #1\strut}}

\newtheorem{theorem}{Theorem}
\newtheorem{lemma}[theorem]{Lemma}

\DeclareMathOperator{\Var}{Var}

\DeclareMathOperator{\logit}{logit}

\usepackage{authblk}

\begin{document}

\title{Gradient tree boosting with random output projections for multi-label classification and
multi-output regression}
% \titlerunning{Random output space projections for gradient boosting }
% \author{Arnaud Joly \and Louis Wehenkel \and Pierre Geurts}
\author[1]{Arnaud Joly}
\author[1]{Louis Wehenkel}
\author[1]{Pierre Geurts}

\affil[1]{Department of EE \& CS, Montefiore Institute\\ University of Liège, Belgium}
% \institute{Department of EE \& CS, Montefiore Institute\\ University of Liège, Belgium \\
% \email{a.joly@ulg.ac.be, l.wehenkel@ulg.ac.be, p.geurts@ulg.ac.be}}

\maketitle

\begin{abstract}
In many applications of supervised learning, multiple classification
or regression outputs have to be predicted jointly. We consider
several extensions of gradient boosting to address such problems. We
first propose a straightforward adaptation of gradient boosting
exploiting multiple output regression trees as base learners. We
then argue that this method is only expected to be optimal when the
outputs are fully correlated, as it forces the partitioning induced
by the tree base learners to be shared by all outputs. We then
propose a novel extension of gradient tree boosting to specifically
address this issue. At each iteration of this new method, a
regression tree structure is grown to fit a single random projection
of the current residuals and the predictions of this tree are fitted
linearly to the current residuals of all the outputs,
independently. Because of this linear fit, the method can adapt
automatically to any output correlation structure. Extensive
experiments are conducted with this method, as well as other
algorithmic variants, on several artificial and real
problems. Randomly projecting the output space is shown to provide a
better adaptation to different output correlation patterns and is
therefore competitive with the best of the other methods in most
settings. Thanks to model sharing, the convergence speed is also
improved, reducing the computing times (or the complexity of the
model) to reach a specific accuracy.

% \keywords{Gradient boosting \and Multi-label learning \and Multi-output regression
% \and Random projection} % disqble

\end{abstract}

\section{Introduction}\label{sec:introduction}

Multi-output supervised learning aims to model input-output relationships
from observations of input-output pairs whenever the output space is a vector
of random variables. Multi-output classification and regression tasks have
numerous applications in domains ranging from biology to multimedia, and recent
applications in this area correspond to very high dimensional
output spaces \cite{agrawal2013multi,dekel2010multiclass}.

Classification and regression trees~\cite{breiman1984classification} are popular
supervised learning methods that provide state-of-the-art performance when
exploited in the context of ensemble methods, namely Random
forests~\cite{breiman2001random,geurts2006extremely} and
Boosting~\cite{freund1997decision,friedman2001greedy}. Classification and
regression trees can obviously be exploited to handle multi-output problems. The
most straightforward way to address multi-output tasks is to apply standard
single output methods separately and independently on each output. Although
simple, this method, called binary relevance~\cite{tsoumakas2009mining} in
multi-label classification or single target~\cite{spyromitros2012multi} in
multi-output regression is often suboptimal as it does not exploit potential
correlations that might exist between the outputs. Tree ensemble methods have
however been explicitely extended by several authors to the joint prediction of
multiple outputs \citep[e.g., ][]{segal1992tree,blockeel2000top}. These extensions
build a single tree to predict all outputs at once. They adapt the score measure
used to assess splits during the tree growth to take into account all outputs
and label each tree leaf with a vector of values, one for each output. Like
standard classification or regression trees, multiple output trees can be
exploited in the context of random
forests~\cite{barutcuoglu2006hierarchical,joly2014random,kocev2007ensembles,kocev2013tree,segal2011multivariate,haider2015copula}
or boosting~\cite{geurts2007gradient} ensembles, which often offer very
significant accuracy improvements with respect to single trees. Multiple output
trees have been shown to be competitive with other multiple output
methods~\cite{madjarov2012extensive}, but, to the best of our knowledge, it has
not been studied as extensively in the context of gradient tree boosting.

Binary relevance / single target of single output tree models and multiple
output tree models represent two extremes in terms of tree structure learning:
the former builds a separate tree ensemble structure for each output, while the
latter builds a single tree ensemble structure for all outputs. Building
separate ensembles for each output may be rather inefficient when the outputs
are strongly correlated. Correlations between the outputs could indeed be
exploited either to reduce model complexity (by sharing the tree structures
between several outputs) or to improve accuracy by regularization. Trying to fit
a single tree structure for all outputs seems however counterproductive when the
outputs are independent. Indeed, in the case of independent outputs,
simultaneously fitting all outputs with a single tree structure may require a
much more complex tree structure than the sum of the individual tree
complexities required to fit the individual outputs. Since training a more
complex tree requires a larger learning sample, multiple output trees are
expected to be outperformed by binary relevance / single target in this
situation.

In this paper, we first formally adapt gradient boosting to multiple output
tasks. We then propose a new method that aims at circumventing the limitations
of both binary relevance / single target and multiple output methods, in the
specific context of tree-based base-learners. Our method is an extension of
gradient tree boosting that can adapt itself to the presence or absence of
correlations between the outputs. At each boosting iteration, a single
regression tree structure is grown to fit a single random projection of the
outputs, or more precisely, of their residuals with respect to the previously
built models. Then, the predictions of this tree are fitted linearly to the
current residuals of all the outputs (independently). New residuals are then
computed taking into account the resulting predictions and the process repeats
itself to fit these new residuals. Because of the linear fit, only the outputs
that are correlated with the random projection at each iteration will benefit
from a reduction of their residuals, while outputs that are independent of the random
projection will remain mostly unaffected. As a consequence, tree structures will
only be shared between correlated outputs as one would expect. Another variant
that we explore consists in replacing the linear global fit by a relabelling of
all tree leaves for each output in turn.

The paper is structured as follows. We show how to extend the gradient
boosting algorithms to multi-output tasks in
Section~\ref{sec:gb-mo-method}. We provide for these algorithms a
convergence proof on the training data and discuss the effect of the
random projection of the output space. We study empirically the
proposed approach in Section~\ref{sec:experiments}. Our first
experiments compare the proposed approaches to binary relevance /
single target on artificial datasets where the output correlation is
known. We also highlight the effect of the choice and size of the
random projection space. We finally carry out an empirical evaluation
of these methods on 21 real-world multi-label and 8 multi-output
regression tasks. Section~\ref{sec:rw} discusses related works and
Section~\ref{sec:conclusions} presents our conclusions.

\section{Background}\label{sec:background}

We denote by $\cal{X}$ an input space, and by $\mathcal{Y}$ an output space;
we suppose that $\mathcal{X} = \mathbb{R}^{p}$ (where $p$ denotes the number of
input features), and that $\mathcal{Y} = \mathbb{R}^{d}$ (where $d$ is the
dimension of the output space). We denote by $P_{{\cal X},{\cal Y}}$ the joint
(unknown) sampling density over $\mathcal{X} \times \mathcal{Y}$.
Given a learning sample $\left((x^i, y^i) \in \left(\mathcal{X} \times
\mathcal{Y}\right)\right)_{i=1}^n$ of $n$ observations in the form of input-output
pairs, a supervised learning task is defined as searching for a function $f^{*} :
\mathcal{X} \rightarrow \mathcal{Y}$ in a hypothesis space $\mathcal{H} \subset
\mathcal{Y}^\mathcal{X}$ that minimizes the expectation of some loss function
$\ell : \mathcal{Y} \times \mathcal{Y} \rightarrow \mathbb{R}$ over the joint
distribution of input / output pairs: $f^{*} \in \arg \min_{f \in \mathcal{H}}
E_{P_{{\cal X},{\cal Y}}} \{ \ell(f(x), y)\}$.

NOTATIONS: Superscript indices ($x^{i}, y^{i}$) denote (input, output) vectors
of an observation $i \in \{1, \ldots , n\}$. Subscript indices (e.g.\ $x_{j},
y_{k}$) denote components of vectors.

We present single and multiple output decision trees in
Section~\ref{subsec:mo-dt}, gradient boosting in
Section~\ref{subsec:gradient-boosting-so} and random projections in
Section~\ref{subsec:random-projection}.

\subsection{Single and multiple output decision trees and forests}
\label{subsec:mo-dt}

A decision tree model~\cite{breiman1984classification} is a
hierarchical set of questions leading to a prediction. The internal
nodes, also called test nodes, test the value of a feature. A test
node $N_t$ has two children called the left child and the right child;
it furthermore has a splitting rule $s_t$ testing whether or not a
sample belongs to its left or right child. For a continuous or
categorical ordered input, the splitting rules are typically of the
form $s_t(x) = x_{F_t} \leq \tau_t$ testing whether or not the input
value $x_{F_t}$ is smaller or equal to a constant $\tau_t$. For a
binary or categorical input, the splitting rule is of the form $s_t(x)
= x_{F_t} \in B_t$ testing whether or not the input value $x_{F_t}$
belongs to the subset of values $B_t$. To predict an unseen sample,
one starts at the root node and follows the tree structure until
reaching a leaf labelled with a prediction $\beta_t \in \mathcal{Y}$.

\paragraph{Random forests.}

Single decision trees are often not very accurate because of a very
high variance. One way to reduce this variance is to grow instead an
ensemble of several randomized trees and then to average their
predictions. For example, \citet{breiman2001random}'s random forests
method builds an ensemble of trees where each tree is trained on a
bootstrap copy of the training set and the best split selection is
randomized by searching this split among $k$ features randomly
selected at each node among the $p$ original features ($k\leq
p$). This randomization can be applied in both single and multi-output
settings
\cite{kocev2007ensembles,segal2011multivariate,joly2014random}. In
what follows, we will compare our gradient boosting approaches with both binary relevance /
single target of single output random forests and with multi-output
random forests.

\subsection{Gradient boosting}
\label{subsec:gradient-boosting-so}

Boosting methods fit additive models of the following form:
\begin{equation}
f(x) = \sum_{m=1}^M \alpha_m f_m(x),
\end{equation}
where $\{f_m\}_{m=1}^M$ are $M$ weak models, so as to minimize some loss
function $\ell:\mathcal{Y} \times \mathcal{Y} \rightarrow \mathbb{R}^+$ over
the learning sample.  This additive model is fit sequentially using a forward
stagewise strategy, adding at each step a new model $f_m(x)$, together with its
weight $\alpha_m$, to the previous $m-1$ models so as to minimize the loss over the
learning sample:
\begin{equation}
\min_{\alpha_m, f_m \in \mathbb{R} \times \mathcal{H}}
\sum_{(x,y) \in \mathcal{L}}
\ell\left(y, \sum_{l=1}^{m-1} \alpha_l f_l(x) + \alpha_m f_m(x)\right),
\label{eq:boosting-optim}
\end{equation}
where $\cal H$ is the hypothesis space of all candidate base functions, e.g.,
the set of all regression trees of some maximum depth. Solving
(\ref{eq:boosting-optim}) is however difficult for general loss
function. Inspired by gradient descent, gradient boosting
\cite{friedman2001greedy} replaces the direct resolution of
(\ref{eq:boosting-optim}) by a single gradient descent step in the functional
space and therefore can be applied for any differentiable loss function. The
whole procedure is detailed in Algorithm
\ref{alg:so-gradient-boosting}. Starting from an initial constant estimate
$\rho_0 \in \mathbb{R}$ (line 2), gradient boosting iteratively follows the
negative gradient of the loss $\ell$ (computed in line 4) as estimated by a
regression model $g_m$ fitted over the training samples using least-square
(line 5) and using an optimal step length $\rho_m$ minimizing the loss $\ell$
(computed in line 6). Independently of the loss, the base model is fitted using
least-square and the algorithm can thus exploit any supervised learning
algorithm that accept square loss to obtain the individual models, e.g.,
standard regression trees.

\begin{algorithm}
\caption{Gradient boosting algorithm}\label{alg:so-gradient-boosting}
\begin{algorithmic}[1]
\Function{GradientBoosting}{$\mathcal{L} = \{(x^i, y^i)\in\mathcal{X}\times\mathcal{Y}\}_{i=1}^n; \ell; \mathcal{H}; M$}
\State $f_0(x) = \rho_0 = \arg\min_{\rho \in \mathbb{R}} \sum_{i=1}^n \ell(y^i, \rho)$.
       \label{alg-line:gb-starting-model}
\For{$m$ = 1 to $M$}
    \State Compute the loss gradient for the training set points
           \[\quad g_m^{i} =
            \left[
            \frac{\partial}{\partial y'}\ell(y^i, y')
            \right]_{y' = f_{m-1}(x)} \forall i \in \left\{1, \ldots, n\right\}.
            \]
            \label{alg-line:boosting-gradient}
    \State Find a correlated direction to the loss gradient
           \[\quad
           g_m = \arg\min_{g \in \mathcal{H}} \sum_{i=1}^n (- g_m^i  - g(x^i))^2.
           \]
    \State Find an optimal step length in the direction $g_m$
            \[
            \quad \rho_m = \arg\min_{\rho \in \mathbb{R}}
              \sum_{i=1}^n \ell\left(y^i, f_{m-1}(x^i) + \rho g_m(x^i)\right).
            \]
          \label{alg-line:gb-step-length}
    \State $f_m(x) = f_{m-1}(x)  + \mu \rho_m g_m(x)$. \label{alg-line:learning_rate}
\EndFor
\State \Return $f_M(x)$
\EndFunction
\end{algorithmic}
\end{algorithm}

The algorithm is completely defined once we have (i)
the constant minimizing the chosen loss (line~\ref{alg-line:boosting-gradient})
and (ii) the gradient of the loss
(line~\ref{alg-line:gb-starting-model}). Table~\ref{tab:common-loss-boosting}
shows these two elements for the square loss in regression and the exponential
loss in classification, as these two losses will be exploited later in our
experiments. With square loss, step 4 actually simply fits the actual residuals
of the problem and the algorithm is called least-squares regression
boosting~\cite{hastie2005elements}. With exponential loss, the algorithm
reduces to the exponential classification boosting algorithm of
\citet{zhu2009multi}. \cite{friedman2001greedy} gives the details how to handle
loss functions such as the absolute loss or the logistic loss.

The optimal step length (line~\ref{alg-line:gb-step-length} of
Algorithm~\ref{alg:so-gradient-boosting}) can be computed either analytically
as for the square loss (see Section~\ref{subsec:output-sampling}) or
numerically using, e.g., \citet{brent2013algorithms}'s method, a robust
root-finding method allowing to minimize single unconstrained optimization
problem, as for the exponential loss. \citet{friedman2001greedy} advises to use
one step of the Newton-Raphson method. However, the Newton-Raphson algorithm
might not converge if the first and second derivatives of the loss are
small. These conditions occur frequently in highly imbalanced supervised
learning tasks.

A learning rate $\mu \in (0, 1]$ is often added
(line~\ref{alg-line:learning_rate}) to shrink the size of the gradient step
$\rho_m$ in the residual space in order to avoid overfitting the learning
sample. Another possible modification is to introduce randomization, e.g. by
subsampling without replacement the samples available (from all learning
samples) at each iteration~\cite{friedman2002stochastic}. In our experiment
with trees, we will introduce randomization by growing the individual
trees with the feature randomization of random forests (parameterized by the
number of features $k$ randomly selected at each node).

\begin{table}
\caption{Constant minimizers and gradient of square loss in regression
  ($\mathcal{Y} = \mathbb{R}$) and exponential loss in binary classification ($\mathcal{Y} = \{-1, 1\}$).}
\label{tab:common-loss-boosting}
\centering
\begin{tabular}{@{}lccc@{}}
\toprule
Loss & $\ell(y, y')$ & $f_0(x)$ & $-\partial{}\ell(y, y')/\partial{}y'$   \\
\midrule
Square  & $\frac{1}{2} (y - y')^2$ & $\frac{1}{n}\sum_{i=1}^n y^i$ & $y - y'$ \\
Logistic & $\log(1+\exp(-2 y y'))$ & $\log\left(\frac{\sum_{i=1}^n 1(y^i=1)}{\sum_{i=1}^n 1(y^i=-1)}\right)$ & $\frac{2y}{1+\exp(2yy')}$\\
\bottomrule
\end{tabular}
\end{table}

\subsection{Random projections}
\label{subsec:random-projection}

Random projection is a very simple and efficient way to reduce the
dimensionality of a high-dimensional space. The general idea of this
technique is to map each data vector $y$ initially of size $d$ into a
new vector $\Phi y$ of size $q$ by multiplying this vector by a matrix
$\Phi$ of size $q\times d$ drawn randomly from some distribution. Since the
projection matrix is random, it can be generated very efficiently. The
relevance of the technique in high-dimensional spaces originates from
the Johnson-Lindenstrauss lemma, which is stated as follows:
\begin{lemma} {\cite{johnson1984extensions}}
\label{lemma:jl-lemma}
Given $\epsilon > 0$ and an integer $n$, let $q$ be a positive integer such
that  $q > 8 \epsilon^{-2} \ln {n}$. For any sample $(y^i)_{i=1}^{n}$ of $n$ points
in $\mathbb{R}^d$ there exists a matrix $\Phi \in \mathbb{R}^{q \times d}$
such that for all $i, j  \in \{1, \ldots , n\}$
\begin{equation}
(1 \hspace*{-0.3mm} - \hspace*{-0.3mm}\epsilon) ||y^i \hspace*{-0.3mm}- \hspace*{-0.3mm}y^j||^2 \leq || \Phi y^i \hspace*{-0.3mm}- \hspace*{-0.3mm}\Phi y^j ||^2
                           \leq (1\hspace*{-0.3mm} +\hspace*{-0.3mm} \epsilon) || y^i \hspace*{-0.3mm}- \hspace*{-0.3mm}y^{j}||^2.
\label{eqn:js}
\end{equation}
\end{lemma}
This lemma shows that when the dimension of the data is high, it is
always possible to map linearly the data into a low-dimensional space
where the original distances are preserved. When $d$ is sufficiently
large, it has been shown that matrices that satisfy
Equation~\ref{eqn:js} with high probability can be generated randomly
in several ways. For example, this is the case of the following two
families of random matrices \cite{achlioptas2003database}:
\begin{itemize}
  \item {\it Gaussian} matrices whose elements are drawn {\em i.i.d.} in
    $\mathcal{N}(0, 1 / q)$;
\item (sparse) {\it Rademacher} matrices whose elements are drawn in the
  finite set $\left\{ -\sqrt{\frac{s}{q}}, 0, \sqrt{\frac{s}{q}}
  \right\}$ with probability $\left\{ \frac{1}{2s}, 1 - \frac{1}{s}
  ,\frac{1}{2s}\right\}$, where $1 / s \in (0,1]$ controls the
    sparsity of $\Phi$.
\end{itemize}
When $s=3$, Rademacher projections will be called {\it Achlioptas}
random projections~\cite{achlioptas2003database}. When the size of the
original space is $p$ and $s=\sqrt{p}$, then we will call the
resulting projections {\it sparse random projections} as
in~\cite{li2006very}. Formal proofs that these random matrices all
satisfy Equation~\ref{eqn:js} with high probability can be found in
the corresponding papers. In all cases, the probability of satisfying
Equation~\ref{eqn:js} increases with the number of random projections
$q$. The choice of $q$ is thus a trade-off between the quality of the
approximation and the size of the resulting embedding.

Note that selecting randomly $q$ dimensions among the original ones is
also a random projection scheme. The corresponding projection
matrix $\Phi$ is obtained by sub-sampling (with or without replacement) q
rows from the $d\times d$ identity
matrix~\cite{candes2011probabilistic}. Later in the paper, we will
call this random projection {\it random output sub-sampling}.

\section{Gradient boosting with multiple outputs}
\label{sec:gb-mo-method}

Starting from a multi-output loss, we first show in
Section~\ref{subsec:gb-extend-mo} how to extend the standard gradient boosting
algorithm to solve multi-output tasks, such as multi-output regression and
multi-label classification, by exploiting existing weak model learners suited
for multi-output prediction. In Section~\ref{subsec:projections}, we then
propose to combine single random projections of the output space with gradient
boosting to automatically adapt to the output correlation structure on these
tasks. This constitutes the main methodological contribution of the paper. We
discuss and compare the effect of the random projection of the output space in
Section~\ref{sec:effect-rp-gb}. Note that we give a convergence proof on the
training data for the proposed algorithms in
Appendix~\ref{sec:convergence-proof}.

\subsection{Standard extension of gradient boosting to multi-output tasks}
\label{subsec:gb-extend-mo}

A loss function $\ell(y,y') \in \mathbb{R}^{+}$ computes the difference between
a ground truth $y$ and a model prediction $y'$. It compares scalars with single
output tasks and vectors with multi-output tasks. The two most common regression
losses are the square loss $\ell_{square}(y, y')= \frac{1}{2} (y-y')^2$ and the
absolute loss $\ell_{absolute}(y, y) = |y - y'|$. Their multi-output extensions
are the $\ell_2$-norm and $\ell_1$-norm losses:
\begin{align}
\ell_2(y, y') &= \frac{1}{2}||y - y'||_{\ell_2}^2,\\
\ell_1(y, y') &= ||y - y'||_{\ell_1}.
\end{align}

In classification, the most commonly used loss to compare a ground truth $y$ to
the model prediction $f(x)$ is the $0-1$ loss $\ell_{0-1}(y,y') = 1(y\not=y')$,
where $1$ is the indicator function. It has two standard multiple output
extensions (i) the Hamming loss $\ell_{Hamming}$ and (ii) the subset $0-1$  loss
$\ell_{\text{subset }0-1}$:
\begin{align}
\ell_{Hamming}(y, y') &= \sum_{j=1}^d 1(y_j \not= y'_{j}), \\
\ell_{\text{subset }0-1} (y, y') &= 1(y\not=y').
\end{align}

Since these losses are discrete, they are
non-differentiable and difficult to optimize. Instead, we propose to extend the
logistic loss $\ell_{logistic}(y,y') =  \log(1 + \exp(-2 y y'))$ used for binary
classification tasks to the multi-label case, as follows:
\begin{equation}
\ell_{logistic}(y, y') = \sum_{j=1}^d \log(1 + \exp(-2 y_j y'_{j})),
\end{equation}
\noindent where we suppose that the $d$ components $y_{j}$ of the target output
vector belong to $\{-1, 1\}$, while the $d$ components $y'_{j}$ of the
predictions may belong to $\mathbb{R}$.

Given a training set  $\mathcal{L} = \left((x^i,y^i) \in \mathcal{X} \times
\mathcal{Y}\right)_{i=1}^n$ and one of these multi-output losses $\ell$, we want to learn
a model $f_{M}$ expressed in the following form
\begin{equation}
f_{M}(x) = \rho_0 + \sum_{m=1}^M \rho_m \odot g_m(x),
\label{eq:pred-mo-gbrt}
\end{equation}
\noindent where the terms $g_{m}$ are selected within a hypothesis space
$\mathcal{H}$ of weak multi-output base-learners, the coefficients $\{\rho_m \in
\mathbb{R}^d\}_{m=0}^M$ are $d$-dimensional vectors highlighting the
contributions of each term $g_m$ to the ensemble, and where the symbol $\odot$
denotes the Hadamard product.  Note that the prediction $f_{M}(x) \in
\mathbb{R}^d$ targets the minimization of the chosen loss $\ell$, but a
transformation might be needed to have a prediction in $\mathcal{Y}$, e.g. we
would apply the $\logit$ function to each output for the multi-output logistic
loss to get a probability estimate of the positive classes.

The gradient boosting method builds such a model in an iterative fashion, as
described in Algorithm~\ref{algo:gb-mo}, and discussed below.

\begin{algorithm}
\caption{Gradient boosting with multi-output regressor weak models.}
\label{algo:gb-mo}
\begin{algorithmic}[1]
\Function{GB-mo}{$\mathcal{L} = \left((x^i, y^i)\right)_{i=1}^n; \ell; \mathcal{H}; M$}
\State $f_{0}(x) = \rho_0 = \arg\min_{\rho \in \mathbb{R}^d} \sum_{i=1}^n \ell(y^i, \rho)$
\For{$m$ = 1 to $M$}
    \State Compute the loss gradient for the learning set samples
            \[g_m^i \in \mathbb{R}^d = \left[\nabla_{y'} \ell(y^i, y') \right]_{y' = f_{m-1}(x^i)}
            \forall i \in \left\{1, \ldots, n\right\}.\]
    \State Fit the negative loss gradient
           \[
           g_m = \arg\min_{g \in \mathcal{H}} \sum_{i=1}^n \left|\left|-g_m^i - g(x^i)\right|\right|_{\ell_2}^2.
           \]
    \State Find an optimal step length in the direction of $g_m$
            \[
            \rho_m = \arg\min_{\rho \in \mathbb{R}^d} \sum_{i=1}^n \ell\left(y^i, f_{m-1}(x^i) + \rho \odot g_m(x^i)\right).
            \]
    \State $f_{m}(x)= f_{m-1}(x)  + \rho_m \odot g_m(x)$
\EndFor
\State \Return $f_{M}(x)$
\EndFunction
\end{algorithmic}
\end{algorithm}

To build the ensemble model, we first initialize it with the constant model
defined by the vector $\rho_0 \in \mathbb{R}^d$ minimizing the multi-output loss
$\ell$ (line 2):
\begin{equation}
\rho_0 = \arg\min_{\rho \in \mathbb{R}^d} \sum_{i=1}^n \ell(y^i, \rho).
\end{equation}

At each subsequent iteration $m$, the multi-output gradient boosting approach
adds a new multi-output weak model $g_m(x)$ with a weight $\rho_m$ to the
current ensemble model by approximating the minimization of the
multi-output loss $\ell$:
\begin{equation}
(\rho_m, g_m) = \arg{}\hspace*{-3mm}\min_{(\rho, g) \in \mathbb{R}^{d}\times \mathcal{H}}
\sum_{i=1}^{n} \ell\left(y^{i}, f_{m-1}(x^i) + \rho \odot g(x^{i})\right).
\label{eq:boosting-exact-fsadm}
\end{equation}

To approximate Equation~\ref{eq:boosting-exact-fsadm}, it first fits a
multi-output weak model $g_m$ to model the negative gradient $g_m^i$
of the multi-output loss $\ell$
\begin{equation}
g_m^i \in\mathbb{R}^d = \left[ \nabla_{y'}  \ell(y^i, y') \right]_{y' = f_{m-1}(x^i)}
\end{equation}
\noindent associated to each sample $i \in \mathcal{L}$ of the training set, by minimizing
the $\ell_2$-loss:
\begin{equation}
g_m = \arg\min_{{g} \in \mathcal{H}} \sum_{i=1}^n \left|\left|-g_m^i - {g}(x^i)\right|\right|_{\ell_2}^2.
\end{equation}

It then computes an optimal step length vector $\rho_m \in \mathbb{R}^d$ in the
direction of the weak model $g_m$ to minimize the multi-output
loss $\ell$:
\begin{equation}
\rho_m = \arg\min_{\rho \in \mathbb{R}^d} \sum_{i=1}^n \ell\left(y^i, f_{m-1}(x^i) + \rho \odot g_m(x^i)\right).
\end{equation}

\subsection{Gradient boosting with output projection} \label{subsec:projections}

Binary relevance / single target of gradient boosting models and gradient
boosting of multi-output models (Algorithm~\ref{algo:gb-mo}) implicitly target
two extreme correlation structures. On the one hand, binary relevance / single
target predicts all outputs independently,  thus assuming that outputs are not
correlated. On the other hand, gradient boosting of multi-output models handles
them all together, thus assuming that they are all correlated. Both approaches
thus exploit the available dataset in a rather biased way. To remove this bias,
we propose a more flexible approach that can adapt itself automatically to the
correlation structure among output variables.

Our idea is that a weak learner used at some step of the gradient boosting
algorithm could be fitted on a single random projection of the output space,
rather than always targeting simultaneously all outputs or always targeting a
single a priori fixed output.

We thus propose to first generate at each iteration of the boosting algorithm
one random projection vector of size $\phi_m \in \mathbb{R}^{1 \times d}$. The
weak learner is then fitted on the projection of the current residuals according
to $\phi_m$ reducing dimensionality from $d$ outputs to a single output. A
weight vector $\rho_m \in \mathbb{R}^d$ is then selected to minimize the
multi-output loss $\ell$. The whole approach is described in
Algorithm~\ref{algo:gbrt-rp}. If the loss is decomposable, non zero components
of the weight vector $\rho_m$ highlight the contribution of the current $m$-th
model to the overall loss decrease. Note that sign flips due to the projection
are taken into account by the additive weights $\rho_m$. A single output
regressor can now handle multi-output tasks through a sequence of single random
projections.

The prediction of an unseen sample $x$ by the model produced by Algorithm~\ref{algo:gbrt-rp}
is now given by
\begin{equation}
f(x) = \rho_0 + \sum_{m=1}^M \rho_m g_m(x),
\end{equation}
\noindent where $\rho_{0}\in\mathbb{R}^d$ is a constant prediction, and the
coefficients $\{\rho_m \in \mathbb{R}^d\}_{m=1}^M$ highlight the contribution of
each model $g_m$ to the ensemble. Note that it is different from
Equation~\ref{eq:pred-mo-gbrt} (no Hadamard product), since here the weak models
$g_{m}$ produce single output predictions.

\begin{algorithm}
\caption{Gradient boosting on randomly projected residual spaces.}
\label{algo:gbrt-rp}
\begin{algorithmic}[1]
\Function{GB-rpo}{$\mathcal{L} = \left((x^i, y^i)\right)_{i=1}^n; \ell; \mathcal{H}; M$}
\State $f_{0}(x) = \rho_0 = \arg\min_{\rho \in \mathbb{R}^d} \sum_{i=1}^n \ell(y^i, \rho)$
\For{$m$ = 1 to $M$}
    \State Compute the loss gradient for the learning set samples
    \[g_m^i \in \mathbb{R}^d = \left[\nabla_{y'} \ell(y^i, y') \right]_{y' = f_{m-1}(x^i)}
    \forall i \in \left\{1, \ldots, n\right\}.\]
    \State Generate a random projection $\phi_m \in \mathbb{R}^{1 \times d}$.
    \State Fit the projected loss gradient
    \[
    g_m = \arg\min_{{g} \in \mathcal{H}} \sum_{i=1}^n  \left(-\phi_m  g_m^i - {g}(x^i)\right)^2.
    \]
    \State Find an optimal step length in the direction of $g_m$.
    \[
    \rho_m = \arg\min_{\rho \in \mathbb{R}^d} \sum_{i=1}^n \ell\left(y^i, f_{m-1}(x^i) + \rho g_m(x^i)\right),
    \] \label{alg-line:rho-optim}
    \State $f_{m}(x) = f_{m-1}(x)  + \rho_m g_m(x)$
\EndFor
\State \Return $f_{M}(x)$
\EndFunction
\end{algorithmic}
\end{algorithm}

Whenever we use decision trees as models, we can grow the tree structure on any
output space and then (re)label it in another one as in~\cite{joly2014random}
by (re)propagating the training samples in the tree structure. This idea of
leaf relabelling could be readily applied to Algorithm~\ref{algo:gbrt-rp}
leading to Algorithm~\ref{algo:gbrt-rp-relabel}. After fitting the decision
tree on the random projection(s) and before optimizing the additive weights
$\rho_m$, we relabel the tree structure in the original residual space (line
7). More precisely, each leaf is labelled by the average unprojected residual
vector of all training examples falling into that leaf.
The predition of an unseen sample is then obtained with
Equation~\ref{eq:pred-mo-gbrt} as for Algorithm~\ref{algo:gb-mo}. We will
investigate whether it is better or not to relabel the decision tree structure
in the experimental section. Note that, because of the relabelling,
Algorithm~\ref{algo:gbrt-rp-relabel} can be straightforwardly used in a
multiple random projection context ($q \geq 1$) using a random projection
matrix $\phi_m \in \mathcal{R}^{q \times d}$. The resulting algorithm with
arbitrary $q$ corresponds to the application to gradient boosting of the idea
explored in~\cite{joly2014random} in the context of random forests (see also
Section \ref{sec:rw}). We will study in Section~\ref{subsec:multiple-rp} and
Section~\ref{sec:exp-rp-size} the effect of the size of the projected space
$q$.

\begin{algorithm}
\caption{Gradient boosting on randomly projected residual spaces
         with relabelled decision trees as weak models.}
\label{algo:gbrt-rp-relabel}
\begin{algorithmic}[1]
\Function{GB-relabel-rpo}{$\mathcal{L} = \left((x^i, y^i)\right)_{i=1}^n; \ell; \mathcal{H}; M; q$}
\State $f_{0}(x) = \rho_0 = \arg\min_{\rho \in \mathbb{R}^d} \sum_{i=1}^n \ell(y^i, \rho)$
\For{$m$ = 1 to $M$}
    \parState{Compute the loss gradient for the learning set samples
            \[g_m^i \in \mathbb{R}^d = \left[\nabla_{y'} \ell(y^i, y') \right]_{y' = f_{m-1}(x^i)}
            \forall i \in \left\{1, \ldots, n\right\}.\]}
    \parState{Generate a random projection $\phi_m \in \mathbb{R}^{q \times d}$.}
    \parState{Fit a single-output tree $g_{m}$ on the projected negative loss gradients
           \[
           g_m = \arg\min_{g \in \mathcal{H}} \sum_{i=1}^n \left|\left|-\phi_m g_m^i - {g}(x^i)\right|\right|_{\ell_2}^2.
           \]}
    \parState{Relabel each leaf of the tree $g_m$ in the original (unprojected) residual space, by
    averaging at each leaf the $g_m^i$ vectors of all examples falling into that leaf.}
    \parState{Find an optimal step length in the direction of $g'_{m}$.}
            \[
            \rho_m = \arg\min_{\rho \in \mathbb{R}^d} \sum_{i=1}^n \ell\left(y^i, f_{m-1}(x^i) + \rho \odot g'_m(x^i)\right).
            \] \label{alg-line:rho-optim2}
    \parState{$f_{m}(x) = f_{m-1}(x)  + \rho_m \odot g'_m(x)$}
\EndFor
\State \Return $f_{M}(x)$
\EndFunction
\end{algorithmic}
\end{algorithm}

To the three presented algorithms, we also add a constant learning rate $\mu \in (0,
1]$ to shrink the size of the gradient step $\rho_m$ in the residual space. Indeed, for
a given weak model space $\mathcal{H}$ and a loss $\ell$, optimizing both the
learning rate $\mu$ and the number of steps $M$ typically  improves
generalization performance.

\subsection{Effect of random projections}
\label{sec:effect-rp-gb}

Randomly projecting the output space in the context of the gradient
boosting approach has two direct consequences: (i) it strongly reduces
the size of the output space, and (ii) it randomly combines several
outputs. We will consider here the different random projection
matrices $\phi \in \mathbb{R}^{q \times d}$ described in
Section~\ref{subsec:random-projection}, i.e., from the sparsest to the
densest projections: random output sub-sampling, Rademacher random
projections (Achlioptas and sparse random projections), and Gaussian
projections.

We discuss in more details the random sub-sampling projection in
Section~\ref{subsec:output-sampling} and the impact of the density of random
projection matrices in Section~\ref{subsec:rp-density}.  We study the benefit to
use more than a single random projection of the output space ($q > 1$) in
Section~\ref{subsec:multiple-rp}.

We also highlight the difference in model representations between tree ensemble
techniques, \emph{i.e.} the gradient tree boosting approaches and the random
forest approaches, in Appendix~\ref{subsec:gb-rp-discussions}.

\subsubsection{$\ell_2$-norm loss and random output sub-sampling}
\label{subsec:output-sampling}

The gradient boosting method has an analytical solution when the loss is the
square loss or its extension the $\ell_2$-norm loss $\ell_2(y, y') =
\frac{1}{2}||y - y'||^2$:
\begin{itemize}

\item The constant model $f_0$ minimizing this loss is the average output value
of the training set given by
\begin{equation}
f_0(x) = \rho_0 = \arg\min_{\rho \in \mathbb{R}^d} \sum_{i=1}^n \frac{1}{2}||y^i - \rho||_{\ell_2}^2
= \frac{1}{n} \sum_{i=1}^n y^i.
\end{equation}

\item The gradient of the $\ell_2$-norm loss for the $i$-th sample is the
difference between the ground truth $y^i$ and the prediction of the ensemble
$f$ at the current step $m$ ($\forall i \in \left\{1, \ldots, n\right\}$):
\begin{align}
g_m^i &= \left[\nabla_{y'} \ell(y^i, y') \right]_{y' = f_{m-1}(x^i)}
 = y^i - f_{m-1}(x^i).
\end{align}

\item Once a new weak estimator $g_m$ has been
fitted on the loss gradient $g_m^i$ or the projected gradient $\phi_m g_m^i$
with or without relabelling, we have to optimize the multiplicative weight
vector $\rho_m$ of the new weak model in the ensemble. For Algorithm~\ref{algo:gb-mo} and Algorithm~\ref{algo:gbrt-rp-relabel} that exploit multi-output weak learners, this amounts to
\begin{align}
\rho_m &= \arg\min_{\rho \in \mathbb{R}^d} \sum_{i=1}^n \frac{1}{2}\left|\left|y^i -  f_m(x^i) - \rho \odot g_m(x^i)\right|\right|^2 \\
&= \arg\min_{\rho \in \mathbb{R}^d} \sum_{i=1}^n \frac{1}{2}\left|\left|g_m^i - \rho \odot g_m(x^i)\right|\right|^2
\end{align}
\noindent
which has the following solution:
\begin{equation}
\rho_{m, j} = \frac{\sum_{i=1}^n g_{m,j}^i g_m(x^i)_j}{\sum_{i=1}^n g_m(x^i)_j} \forall j \in \{1,\ldots,d\}.
\label{eq:loss-weight-sub}
\end{equation}

For Algorithm~\ref{algo:gbrt-rp}, we have to solve
\begin{align}
\rho_m &= \arg\min_{\rho \in \mathbb{R}^d} \sum_{i=1}^n \frac{1}{2}\left|\left|y^i -  f_m(x^i) - \rho g_m(x^i)\right|\right|^2 \\
&= \arg\min_{\rho \in \mathbb{R}^d} \sum_{i=1}^n \frac{1}{2}\left|\left|g_m^i - \rho g_m(x^i)\right|\right|^2
\end{align}
\noindent which has the following solution
\begin{equation}
\rho_{m, j} = \frac{\sum_{i=1}^n g_{m,j}^i g_m(x^i)}{\sum_{i=1}^n g_m(x^i)} \forall j \in \{1,\ldots,d\}.
\label{eq:loss-weight-sub-2}
\end{equation}

\end{itemize}

From Equation~\ref{eq:loss-weight-sub} and Equation~\ref{eq:loss-weight-sub-2},
we have that the weight $\rho_{m, j}$ is proportional to the correlation
between the loss gradient of the output $j$ and the weak estimator $g_m$. If
the model $g_m$ is independent of the output $j$, the weight $\rho_{m,j}$
will be close to zero and $g_m$ will thus not contribute to the prediction of
this output. On the opposite, a high magnitude of $|\rho_{m,j}|$ means that the
model $g_m$ is useful to predict the output $j$.

If we subsample the output space at each boosting iteration
(Algorithm~\ref{algo:gbrt-rp} with random output sub-sampling), the weight
$\rho_{m, j}$ is then proportional to the correlation between the model fitted
on the sub-sampled output and the output $j$. If correlations exist between the
outputs, the optimization of the constant $\rho_m$ allows to share the trained
model at the $m$-th iteration on the sub-sampled output to all the other
outputs. In the extreme case where all outputs are independent given the
inputs, the weight $\rho_m$ is expected to be nearly zero for all outputs
except for the sub-sampled one, and Algorithm~\ref{algo:gbrt-rp} would be
equivalent to the binary relevance / single target approach. If all outputs are
strictly identical, the elements of the constant vector $\rho_m$ would have the
same value, and Algorithm~\ref{algo:gbrt-rp} would be equivalent to the
multi-output gradient boosting approach
(Algorithm~\ref{algo:gb-mo}). Algorithm~\ref{algo:gbrt-rp} would also produce
in this case the exact same model as the binary relevance / single target approach
asymptotically but it would require $d$ times less trees to reach similar
performance, as each tree would be shared by all $d$ outputs.

Algorithm~\ref{algo:gbrt-rp-relabel} with random output sub-sampling is a gradient
boosting approach fitting one decision tree at each iteration on a random output
space and relabelling the tree in the original output space. The leaf
relabelling procedure minimizes the $\ell_2$-norm loss over the training
samples by averaging the output values of the samples reaching the corresponding
leaves. In this case, the optimization of the weight $\rho_m$ is unnecessary, as it would
lead to an all ones vector. For similar reasons if the multi-output gradient
boosting method (Algorithm~\ref{algo:gb-mo}) uses decision trees as weak
estimators, the weight $\rho_m$ is also an all ones vector as the leaf
predictions already minimize the $\ell_2$-norm loss. The difference between
these two algorithms is that Algorithm~\ref{algo:gbrt-rp-relabel} grows trees
using a random output at each iteration instead of all of them with
Algorithm~\ref{algo:gb-mo}.

\subsubsection{Density of the random projections}
\label{subsec:rp-density}

\citet{joly2014random} have combined the random forest method with a wide
variety of random projection schemes. While the algorithms presented in this
paper were originally devised with random output sub-sampling in mind (see
Section~\ref{subsec:output-sampling}), it seems natural to also combine the
proposed approaches with random projection schemes such as Gaussian random
projections or (sparse) Rademacher random projections.

With random output sub-sampling, the projection matrix $\phi_m \in
\mathbb{R}^{1 \times d}$ is extremely sparse as only one element is non zero.
With denser random projections, the weak estimators of
Algorithm~\ref{algo:gbrt-rp} and Algorithm~\ref{algo:gbrt-rp-relabel} are fitted
on the projected gradient loss $\{(x^i, \phi_m g_m^i\}_{i=1}^n$. It means that
a weak estimator $g_m$ is trying to model the direction of a weighted
combination of the gradient loss.

Otherwise said, the weak model fitted at the $m$-th step approximates a
projection $\phi_m$ of the gradient losses given the input vector. We can
interpret the weight $\rho_{m,j}$ when minimizing the $\ell_2$-norm loss as the
correlation between the output $j$ and a weighted approximation of the output
variables $\phi_m$. With an extremely sparse projection having only one non zero
element, we have the situation described in the previous section. If we have two
non zero elements, we have the following extreme cases: (i) both combined
outputs are identical and (ii) both combined outputs are independent given the
inputs. In the first situation, the effect is identical to the case where we
sub-sample only one output. In the second situation, the weak model makes a
compromise between the independent outputs given by $\phi_m$. Between those two
extremes, the loss gradient direction $\phi_m$ approximated by the weak model is
useful to predict both outputs. The random projection of the output space will
indeed prevent over-fitting by inducing some variance in the learning process.
The previous reasoning can be extended to more than two output variables.

Dense random projection schemes, such as Gaussian random projection, consider a
higher number of outputs together and is hoped to speed up convergence by
increasing the correlation between the fitted tree in the projected space and
the residual space. Conversely, sparse random projections, such as random output
sub-sampling, make the weak model focus on few outputs.

\subsubsection{Gradient tree boosting and multiple random projections}
\label{subsec:multiple-rp}

The gradient boosting multi-output strategy combining random projections and tree
relabelling (Algorithm~\ref{algo:gbrt-rp-relabel}) can use random projection
matrices $\phi_m \in \mathbb{R}^{q \times d}$ with more than one line
($q\geq 1$).

The weak estimators are multi-output regression trees using the variance as
impurity criterion to grow their tree structures. With an increasing number of
projections $q$, we have the theoretical guarantee that the variance computed in
the projected space is a good approximation of the variance in the original output
space.

\begin{theorem}{\bf Variance preservation with random projection~\cite{joly2014random}}
\label{thm:var-jl-lemma}
Given $\epsilon > 0$, a sample $(y^i)_{i=1}^{n}$ of $n$ points $y \in
\mathbb{R}^d$, and a projection matrix $\Phi\in \mathbb{R}^{q\times d}$ such
that for all $i, j  \in \{1, \ldots , n\}$ the condition given by
Equation~\ref{eqn:js} holds, we have also:
\begin{equation}
(1 - \epsilon) \Var((y^i)_{i=1}^n)  \leq \Var((\Phi y^i)_{i=1}^n)
                                    \leq (1 + \epsilon) \Var((y^i)_{i=1}^n).
\label{eq:var-ineq}
\end{equation}
\end{theorem}

When the projected space is of infinite size $q\rightarrow\infty$, the decision
trees grown on the original space or on the projected space are identical as the
approximation of the variance is exact. We thus have that
Algorithm~\ref{algo:gbrt-rp-relabel} is equivalent to the gradient boosting with
multi-output regression tree method (Algorithm~\ref{algo:gb-mo}).

Whenever the projected space is of finite size ($q < \infty$),
Algorithm~\ref{algo:gbrt-rp-relabel} is thus an approximation of
Algorithm~\ref{algo:gb-mo}. We study empirically the effect of the number of
projections $q$ in Algorithm~\ref{algo:gbrt-rp-relabel} in
Section~\ref{sec:exp-rp-size}.

\section{Experiments}
\label{sec:experiments}

We describe the experimental protocol and real world datasets in
Section~\ref{sec:experimental-protocol}. Our first experiments in
Section~\ref{subsec:synthetic-exp} illustrate the multi-output gradient boosting
methods on synthetic datasets where the output correlation structure is known.
The effect of the nature of random projections of the output
space is later studied for Algorithm~\ref{algo:gbrt-rp} and
Algorithm~\ref{algo:gbrt-rp-relabel} in Section~\ref{sec:gb-exp-rp-effect}. We
compare multi-output gradient boosting approaches and multi-output random forest
approaches in Section~\ref{sec:systematic-analysis} over 29 real multi-label and
multi-output datasets.

\subsection{Experimental protocol and real world datasets}
\label{sec:experimental-protocol}

We describe the metrics used to assess the performance of the supervised
learning algorithms in Section~\ref{subsec:metrics-protocol}. The protocol used
to optimize hyper-parameters is given in
Section~\ref{subsec:hyperparam-protocol}. The real world datasets are
presented in Section~\ref{subsec:real-datasets}.

\subsubsection{Accuracy assessment protocol}
\label{subsec:metrics-protocol}

We assess the accuracy of the predictors for multi-label classification on a
test sample (TS) by the  ``Label Ranking Average Precision
(LRAP)''~\cite{schapire2000boostexter}. For each sample $y^i$, it averages over
each true label $j$ the ratio between (i) the number of true label (i.e.
$y^i=1$) with higher scores or probabilities than the label $j$ to (ii) the
number of labels ($y^i$) with higher score $f(x^i)$ than the label $j$.
Mathematically, we average the LRAP of all pairs of ground truth $y^i$ and its
associated prediction $f(x^i)$:
\begin{equation}
\text{LRAP}(\hat{f})
= \frac{1}{|TS|} \sum_{i=1}^n \frac{1}{|y^i|} \hspace*{-1mm}\sum_{j \in \{k : y^i_{k}=1\}} \hspace*{-1mm}
\frac{|\mathcal{L}_j^{i}(y^i)|}{|\mathcal{L}_j^{i}(1_d)|},
\end{equation}
\noindent where $$\mathcal{L}^i_{j}(q) = \left\{ k :
q_{k}=1 \mbox{~and~} \hat{f}(x^i)_k \geq \hat{f}(x^i)_j\right\}.$$ The best
possible average precision is thus 1. The LRAP score is equal to the fraction of
positive labels if all labels are predicted with the same score or all negative
labels have a score higher than the positive one. Notice that we use
indifferently the notation $|\cdot |$ to express the cardinality of a set or the
$1$-norm of a vector.

We assess the accuracy of the predictors for multi-output regression task
by the ``macro-$r^2$ score'', expressed by
\begin{equation}
\text{macro-}r^2(\hat{f}) = 1 -
\frac{1}{d} \sum_{j=1}^d
\frac{\sum_{i=1}^n (y^i_j - \hat{f}(x^i)_j)^2}{\sum_{i=1}^n (y^i_j - \frac{1}{n}\sum_{l=1}^n y^i_j)^2}
\end{equation}
\noindent where $\hat{f}(x^i)_j$ is the predicted value to the output $j$ by the
learnt model $\hat{f}$ applied to $x^i$. The best possible macro-$r^2$ score is
1. A value below or equal to zero ($\text{macro-}r^2(\hat{f}) \leq 0$) indicates
that the models is worse than a constant.

\subsubsection{Hyper-parameter optimization protocol}
\label{subsec:hyperparam-protocol}

The hyper-parameters of the supervised learning algorithms are optimized as
follows: we define an hyper-parameter grid and the best hyper-parameter set is
selected using $20\%$ of the training samples as a validation set. The results
shown are averaged over five random split of the dataset
while preserving the training-testing set size ratio.

For the boosting ensembles, we optimize the learning rate $\mu$ among $\{1,
0.5, 0.2, 0.1,$ $0.05, 0.02, 0.01\}$ and use decision trees as weak models whose
hyper-parameters are also optimized: the number of features drawn at each node
$k$ during the tree growth is selected among $k \in \{\sqrt{p}, 0.1p, 0.2p,
0.5p, p\}$, the maximum number of tree leaves $n_{\max\_leaves}$ grown in
best-first fashion is chosen among $\{2,\ldots,8\}$. Note
that a decision tree with $n_{\max\_leaves}=2$ and $k=p$ is called a stump. We
add new weak models to the ensemble by minimizing either the square loss or the
absolute loss (or their multi-output extensions) in regression and either the
square loss or the logistic loss (or their multi-output extensions) in
classification, the choice of the loss being an additional hyper-parameter
tuned on the validation set.

We also optimize the number of boosting steps $n_{iter}$ of each gradient
boosting algorithm over the validation set. However note that the number of
steps has a different meaning depending on the algorithm. For binary relevance /
single target gradient boosting, the number of boosting steps $n_{iter}$ gives
the number of weak models fitted per output. The implemented algorithm here fits
weak models in a round robin fashion over all outputs. For all other (multi-output) methods,
the number of boosting steps $n_{iter}$ is the total number of weak models for
all outputs as only one model is needed to fit all outputs. The computing time
of one boosting iteration is thus different between the approaches. We will set
the budget, the maximal number of boosting steps $n_{iter}$, for each algorithm
to $n_{iter}=10000$ on synthetic experiments (see
Section~\ref{subsec:synthetic-exp}) so that the performance of the estimator is
not limited by the computational power. On the real datasets however, this
setting would have been too costly. We decided instead to limit the computing
time allocated to each gradient boosting algorithm on each classification
(resp. regression) problem to $100\times T$ (resp. $500\times T$), where $T$ is
the time needed on this specific problem for one iteration of multi-output
gradient boosting (Algorithm~\ref{algo:gb-mo}) with stumps and the
$\ell_2$-norm loss. The maximum number of iterations, $n_{iter}$, is thus set
independently for each problem and each hyper-parameter setting such that this
time constraint is satisfied. As a consequence, all approaches thus receive
approximately the same global time budget for model training and
hyper-parameter optimization.

For the random forest algorithms, we use the default hyper-parameter setting
suggested in~\cite{hastie2005elements}, which corresponds in classification
to 100 totally developed trees with $k=\sqrt{p}$ and in regression to $100$
trees with $k=p/3$ and a minimum of 5 samples to split a node ($n_{\min{}}=5$).

The base learner implementations are based on the
random-output-trees\footnote{\url{https://github.com/arjoly/random-output-trees}}~\cite{joly2014random}
version 0.1 and on the scikit-learn~\cite{buitinck2013api,pedregosa2011scikit}
of version 0.16 Python package. The algorithms presented in this paper will
be provided in random-output-trees version 0.2.

\subsubsection{Real world datasets}
\label{subsec:real-datasets}

Experiments are conducted on 29 real world multi-label and
multi-output datasets related to various domains ranging from text to
biology and multimedia. The basic characteristics of each dataset are
summarized in Table~\ref{tab:dataset-multi-label} for multi-label
datasets and in Table~\ref{tab:dataset-multi-output regression} for
multi-output regression datasets. For more information on a particular
dataset, please see Appendix~\ref{appendix:real-datasets}. If the
number of testing samples is unspecified, we use $40\%$ of the samples
as training, $10\%$ of the samples as validation set and $50\%$ of the
samples as testing set.

\begin{table}[t]
\caption{Selected multi-label ranging from $d=6$ to $d=1862$ outputs.}
\label{tab:dataset-multi-label}
\centering
\begin{tabular}{lrlrr}
\toprule
            Datasets &  $n_{LS}$ & $n_{TS}$ &    $p$ &   $d$ \\
\midrule
              CAL500 &       502 &          &     68 &   174 \\
              bibtex &      4880 &     2515 &   1836 &   159 \\
               birds &       322 &      323 &    260 &    19 \\
           bookmarks &     87856 &          &   2150 &   208 \\
             corel5k &      4500 &      500 &    499 &   374 \\
           delicious &     12920 &     3185 &    500 &   983 \\
             diatoms &      2065 &     1054 &    371 &   359 \\
    drug-interaction &      1862 &          &    660 &  1554 \\
            emotions &       391 &      202 &     72 &     6 \\
               enron &      1123 &      579 &   1001 &    53 \\
             genbase &       463 &      199 &   1186 &    27 \\
           mediamill &     30993 &    12914 &    120 &   101 \\
             medical &       333 &      645 &   1449 &    45 \\
 protein-interaction &      1554 &          &    876 &  1862 \\
             reuters &      2500 &     5000 &  19769 &    34 \\
               scene &      1211 &     1196 &    294 &     6 \\
             scop-go &      6507 &     3336 &   2003 &   465 \\
     sequence-funcat &      2455 &     1264 &   4450 &    31 \\
                wipo &      1352 &      358 &  74435 &   187 \\
               yeast &      2417 &          &    103 &    14 \\
            yeast-go &      2310 &     1155 &   5930 &   132 \\
\bottomrule
\end{tabular}
\end{table}

\begin{table}[t]
\caption{Selected multi-output regression ranging from $d=2$ to $d=16$ outputs.}
\label{tab:dataset-multi-output regression}
\centering
\begin{tabular}{lrlrr}
\toprule
      Datasets &  $n_{LS}$ & $n_{TS}$ &  $p$ &  $d$ \\
\midrule
         atp1d &       337 &          &  411 &    6 \\
         atp7d &       296 &          &  411 &    6 \\
           edm &       154 &          &   16 &    2 \\
         oes10 &       403 &          &  298 &   16 \\
         oes97 &       334 &          &  263 &   16 \\
         scm1d &      8145 &     1658 &  280 &   16 \\
        scm20d &      7463 &     1503 &   61 &   16 \\
 water-quality &      1060 &          &   16 &   14 \\
\bottomrule
\end{tabular}
\end{table}

\subsection{Experiments on synthetic datasets with known output correlation structures}
\label{subsec:synthetic-exp}

To give a first illustration of the methods, we study here the proposed
boosting approaches on synthetic datasets whose output correlation
structures are known. The datasets are first presented in
Section~\ref{sec:artificial-datasets}. We then compare on these
datasets multi-output gradient boosting approaches in terms of their
convergence speed in Section~\ref{sec:convergence-artificial} and in
terms of their best performance whenever hyper-parameters are
optimized in Section~\ref{sec:synthetic-exp}.

\subsubsection{Synthetic datasets}
\label{sec:artificial-datasets}

We use three synthetic datasets with a specific output structure: (i)
chained outputs, (ii) totally correlated outputs and (iii) fully
independent outputs. Those tasks are derived from the
\textbf{friedman1} regression dataset which consists in solving the
following single target regression
task~\cite{friedman1991multivariate}
\begin{align}
f(x) &= 10 \sin(\pi x_1 x_2)  + 20 (x_3 - 0.5)^2 + 10 x_4 + 5 x_5 \\
y &= f(x) + \epsilon
\end{align}
\noindent with  $x \in \mathbb{R}^5 \sim \mathcal{N}(0; I_5)$ and $\epsilon \sim \mathcal{N}(0;
1)$ where $I_5$ is an identity matrix of size $5 \times 5$.

The \textbf{friedman1-chain} problem consists in $d$ regression tasks
forming a chain obtained by cumulatively adding independent standard
Normal noise. We draw samples from the following distribution
\begin{align}
y_1 &= f(x) + \epsilon_1,\\
y_j &= y_{j-1} + \epsilon_j \quad \forall j \in \{2, \ldots, d \}
\end{align}
\noindent with $x \sim \mathcal{N}(0; I_5)$ and $\epsilon \sim \mathcal{N}(0;
I_{d})$. Given the chain structure, the output with the least amount of noise
is the first one of the chain and averaging a subset of the outputs would not
lead to any reduction of the output noise with respect to the first output,
since total noise variance accumulates more than linearly with the number of
outputs. The optimal multi-output strategy is thus to build a model using only
the first output and then to replicate the prediction of this model for all
other outputs.

The \textbf{friedman1-group} problem consists in solving $d$ regression tasks
simultaneously obtained from one friedman1 problem without noise where an
independent normal noise is added. Given $x \sim \mathcal{N}(0; I_5)$ and
$\epsilon \sim \mathcal{N}(0; I_{d})$, we have to solve the following task:
\begin{align}
y_j =& f(x) + \epsilon_j  \quad \forall j \in \{1, \ldots, d \}.
\end{align}
\noindent If the output-output structure is known, the additive noises
$\epsilon_j, \forall j \in \{1, \ldots, d \},$ can be filtered out by averaging
all outputs. The optimal strategy to address this problem is thus to train a
single output regression model to fit the average output. Predictions on unseen
data would be later done by replicating the output of this model for all
outputs.

The \textbf{friedman1-ind} problem consists in $d$ independent friedman1 tasks.
Drawing samples from  $x \sim \mathcal{N}(0; I_{5 d})$ and $\epsilon \sim
\mathcal{N}(0; I_{d})$, we have
\begin{align}
y_j =& f(x_{5j+1:5j+5}) + \epsilon_j \quad \forall j \in \{1, \ldots, d \}.
\end{align}
\noindent where $x_{5j+1:5j+5}$ is a slice of feature vector from feature $5j+1$
to $5j+5$. Since all outputs are independent, the best multi-output strategy
is single target: one independent model fits each output.

For each multi-output friedman problem, we consider 300 training samples, 4000
testing samples and $d=16$ outputs.

\subsubsection{Convergence with known output correlation structure}
\label{sec:convergence-artificial}

We first study the macro-$r^2$ score convergence as a function of time (see
Figure~\ref{fig:gbrt-subsample}) for three multi-output gradient boosting
strategies: (i) single target of gradient tree boosting (st-gbrt), (ii)
gradient boosting with multi-output regression tree (gbmort,
Algorithm~\ref{algo:gb-mo}) and (iii) gradient boosting with output subsampling
of the output space (gbrt-rpo-subsample, Algorithm~\ref{algo:gbrt-rp}). We train
each boosting algorithm on the three friedman1 artificial datasets with the same
set of hyper-parameters: a learning rate of $\mu=0.1$  and stumps as weak
estimators (a decision tree with $k=p$, $n_{\max{}\_leaves}=2$) while minimizing
the square loss.

% python analyse_plot_time.py -d friedman1_chain_v2_n_out_16_noise_1.0 friedman1_mo_v3_16_1_16 friedman1_mo_16_16_1 -t subsampled  --dir subsample  --relabel False
% python analyse_plot_time.py -d friedman1_chain_v2_n_out_8_noise_1.0_noisy-out friedman1_mo_v3_8_1_8_noisy-out friedman1_mo_8_8_1_noisy-out -t subsampled  --dir noisy-out  --relabel False

\begin{figure}
\centering
\subfloat[Friedman1-chain]{\label{subfig:subsample-chain}\includegraphics[width=0.7\textwidth]{{{friedman-subs_chain_v2_n_out_16_noise_1.0_log_time_macro-r2}}}} \\
\subfloat[Friedman1-group]{\label{subfig:subsample-gr}\includegraphics[width=0.7\textwidth]{{{friedman-subs_mo_v3_16_1_16_log_time_macro-r2}}}}\\
\subfloat[Friedman1-ind]{\label{subfig:subsample-ind}\includegraphics[width=0.7\textwidth]{{{friedman-subs_mo_16_16_1_log_time_macro-r2}}}}
\caption{The convergence speed and the optimum reached are affected by the
output correlation structure. The gbmort and gbrt-rpo-subsampled algorithms both exploit
output correlations, which yields faster convergence and slightly better performance than
st-gbrt on friedman1-chain and friedman1-group. However, st-gbrt
converges to a better optimum than gbmort and gbrt-rpo-subsample when there is no
output correlation as in friedman1-ind. (Model parameters: $k=p$,
$n_{\max{}\_leaves}=2$, $\mu=0.1$)}
\label{fig:gbrt-subsample}
\end{figure}

On the friedman1-chain (see Figure~\ref{subfig:subsample-chain}) and
friedman1-group (see Figure~\ref{subfig:subsample-gr}), gbmort and
gbrt-rpo-subsampled converge more than 100 times faster (note the logarithmic
scale of the abscissa) than single target. Furthermore, the optimal
macro-$r^2$ is slightly better for gbmort and gbrt-rpo-subsampled than st-gbrt.
All methods are over-fitting after a sufficient amount of time.
Since all outputs are correlated on both datasets, gbmort and
gbrt-rpo-susbsampled are exploiting the output structure to have faster
convergence. The gbmort method exploits the output structure by filtering the
output noise as the stump fitted at each iteration is the one that maximizes the
reduction of the average output variance. By contrast, gbrt-rpo-subsample
detects output correlations by optimizing the $\rho_m$ constant and then shares
the information obtained by the current weak model with all other outputs.

On the friedman1-ind dataset (see Figure~\ref{subfig:subsample-ind}),
all three methods converge at the same rate. However, the single target
strategy converges to a better optimum than gbmort and gbrt-rpo-subsample. Since
all outputs are independent, single target enforces the proper correlation
structure (see Figure~\ref{subfig:subsample-ind}). The gbmort method has the
worst performance as it assumes the wrong set of hypotheses. The
gbrt-rpo-subsampled method pays the price of its flexibility by over-fitting the
additive weight associated to each output, but less than gbmort.

This experiment confirms that enforcing the right correlation structure yields
faster convergence and the best accuracy. Nevertheless, the output structure is
unknown in practice. We need flexible approaches such as gbrt-rpo-subsampled
that automatically detects and exploits the correlation structure.

\subsubsection{Performance and output modeling assumption}
\label{sec:synthetic-exp}

The presence or absence of structures among the outputs has shown to affect the
convergence speed of multi-output gradient boosting methods. As discussed
in~\cite{dembczynski2012label}, we talk about conditionally independent outputs when:
$$P(y_1,\ldots,y_q|x)=P(y_1|x)\cdots P(y_q|x)$$ and about unconditionally
independent outputs when:$$P(y_1,\ldots,y_q)=P(y_1)\cdots P(y_q).$$ When the
outputs are not conditionally independent and the loss function can not be
decomposed over the outputs (eg., the subset $0-1$ loss), one might need to
model the joint output distribution $P(y_1,\ldots,y_q|x)$ to obtain a Bayes
optimal prediction. If the outputs are conditionally independent however or if
the loss function can be decomposed over the outputs, then a Bayes optimal
prediction can be obtained by modeling separately the marginal conditional
output distributions $P(y_j|x)$ for all $j$. This suggests that in this case,
binary relevance / single target is not really penalized asymptotically with
respect to multiple output methods for not considering the outputs jointly. In
the case of an infinite sample size, it is thus expected to provide as good
models as all the multiple output methods. \emph{Since in practice we have to
deal with finite sample sizes, multiple output methods may provide better
results by better controlling the bias/variance trade-off.}

Let us study this question on the three synthetic datasets: friedman1-chain,
friedman1-group or friedman1-ind. We optimize the hyper-parameters with a
computational budget of 10000 weak models per hyper-parameter set. Five
strategies are compared (i) the artificial-gbrt method, which assumes that the
output structure is known and implements the optimal strategy on each problem as
explained in Section~\ref{sec:artificial-datasets}, (ii) single target of
gradient boosting regression trees (st-gbrt), (iii) gradient boosting with
multi-output regression tree (gbmort, Algorithm~\ref{algo:gb-mo}) and gradient
boosting with randomly sub-sampled outputs (iv) without relabelling
(gbrt-rpo-subsampled, Algorithm~\ref{algo:gbrt-rp}) and (v) with relabelling
(gbrt-relabel-rpo-subsampled, Algorithm~\ref{algo:gbrt-rp-relabel}). All
boosting algorithms minimize the square loss, the absolute loss or their
multi-outputs extension the $\ell_2$-norm loss.

We give the performance on the three tasks for each estimator in
Table~\ref{table:friedman1} and the p-value of Student's paired $t$-test
comparing the performance of two estimators on the same dataset in
Table~\ref{tab:tstudent-friedman1}.

% python analyse.py -d friedman1_chain_v2_n_out_16_noise_1.0 friedman1_mo_v3_16_1_16 friedman1_mo_16_16_1  --to-latex

\begin{table}
\caption{All  methods compared on the 3 artificial datasets. Exploiting the
output correlation structure (if it exists) allows beating single target
in a finite sample size, decomposable metric and conditionally independent
output.}\small
\label{table:friedman1}
\centering
\begin{tabular}{llll}
\toprule
Dataset &          friedman1-chain &        friedman1-group &          friedman1-ind \\
\midrule
artificial-gbrt             &  $0.654 \pm 0.015$ (1) &  $0.889 \pm 0.009$ (1) &  $0.831 \pm 0.004$ (1) \\
st-gbrt                     &  $0.626 \pm 0.016$ (5) &  $0.873 \pm 0.008$ (5) &  $0.830 \pm 0.003$ (2) \\
gbmort                      &  $0.640 \pm 0.008$ (4) &  $0.874 \pm 0.012$ (4) &  $0.644 \pm 0.010$ (5) \\
gbrt-relabel-rpo-subsampled &  $0.648 \pm 0.015$ (2) &  $0.880 \pm 0.009$ (2) &  $0.706 \pm 0.009$ (4) \\
gbrt-rpo-subsampled         &  $0.645 \pm 0.013$ (3) &  $0.876 \pm 0.007$ (3) &  $0.789 \pm 0.003$ (3) \\
\bottomrule
\end{tabular}
\end{table}

\begin{table}
\caption{P-values given by Student's paired $t$-test on the synthetic datasets.
We highlight p-values inferior to $\alpha=0.05$ in bold. Note that the sign $>$
(resp. $<$) indicates that the estimator in the row has better (resp.lower)
score than the column estimator.}
\label{tab:tstudent-friedman1}
\centering
% \begin{small}
\begin{tabular}{llllll}
% \toprule
{} & \rotatebox{90}{artificial-gbrt} & \rotatebox{90}{st-gbrt} & \rotatebox{90}{gbmort} & \rotatebox{90}{gbrt-relabel-rpo-subsampled} & \rotatebox{90}{gbrt-rpo-subsampled} \\
\midrule
Dataset friedman1-chain \\
\midrule
artificial-gbrt             &                                 &       \textbf{0.003}(>) &                   0.16 &                                        0.34 &                                0.24 \\
st-gbrt                     &               \textbf{0.003}(<) &                         &                   0.11 &                            \textbf{0.04}(<) &                    \textbf{0.03}(<) \\
gbmort                      &                            0.16 &                    0.11 &                        &                                        0.38 &                                0.46 \\
gbrt-relabel-rpo-subsampled &                            0.34 &        \textbf{0.04}(>) &                   0.38 &                                             &                                0.57 \\
gbrt-rpo-subsampled         &                            0.24 &        \textbf{0.03}(>) &                   0.46 &                                        0.57 &                                     \\
\midrule
Dataset friedman1-group \\
\midrule
artificial-gbrt             &                                 &       \textbf{0.005}(>) &      \textbf{0.009}(>) &                           \textbf{0.047}(>) &                   \textbf{0.006}(>) \\
st-gbrt                     &               \textbf{0.005}(<) &                         &                   0.56 &                           \textbf{0.046}(<) &                                0.17 \\
gbmort                      &               \textbf{0.009}(<) &                    0.56 &                        &                                        0.15 &                                0.63 \\
gbrt-relabel-rpo-subsampled &               \textbf{0.047}(<) &       \textbf{0.046}(>) &                   0.15 &                                             &                    \textbf{0.04}(>) \\
gbrt-rpo-subsampled         &               \textbf{0.006}(<) &                    0.17 &                   0.63 &                            \textbf{0.04}(<) &                                     \\
\midrule
Dataset friedman1-ind \\
\midrule
artificial-gbrt             &                                 &                    0.17 &      \textbf{2e-06}(>) &                           \textbf{2e-06}(>) &                   \textbf{1e-05}(>) \\
st-gbrt                     &                            0.17 &                         &      \textbf{2e-06}(>) &                           \textbf{4e-06}(>) &                   \textbf{4e-06}(>) \\
gbmort                      &               \textbf{2e-06}(<) &       \textbf{2e-06}(<) &                        &                           \textbf{9e-05}(<) &                   \textbf{6e-06}(<) \\
gbrt-relabel-rpo-subsampled &               \textbf{2e-06}(<) &       \textbf{4e-06}(<) &      \textbf{9e-05}(>) &                                             &                   \textbf{3e-05}(<) \\
gbrt-rpo-subsampled         &               \textbf{1e-05}(<) &       \textbf{4e-06}(<) &      \textbf{6e-06}(>) &                           \textbf{3e-05}(>) &                                     \\
\bottomrule
\end{tabular}
% \end{small}
\end{table}

As expected, we obtain the best performance if the output correlation structure
is known with the custom strategies implemented with artificial-gbrt. Excluding
this artificial method, the best boosting methods on the two problems with
output correlations, friedman1-chain and friedman1-group, are the two gradient
boosting approaches with output subsampling (gbrt-relabel-rpo-subsampled and
gbrt-rpo-subsampled).

In friedman1-chain, the output correlation structure forms a chain as each new
output is the previous one in the chain with a noisy output. Predicting outputs
at the end of the chain, without using the previous ones, is a difficult task.
The single target approach is thus expected to be sub-optimal. And indeed, on
this problem, artificial-gbrt, gbrt-relabel-rpo-subsampled and
gbrt-rpo-subsampled are significantly better than st-gbrt (with
$\alpha=0.05$). All the multi-output methods, including gbmort, are
indistinguishable from a statistical point of view, but we note that gbmort is
however not significantly better than st-gbrt.

In friedman1-group, among the ten pairs of algorithms, four are not
significantly different, showing a p-value greater than 0.05 (see
Table~\ref{tab:tstudent-friedman1}). We first note that gbmort is not better
than st-gbrt while exploiting the correlation. Secondly, the boosting methods
with random output sub-sampling are the best methods. They are however not
significantly better than gbmort and significantly worse than artificial-gbrt,
which assumes the output structure is known. Note that gbrt-relabel-rpo-subsampled
is significantly better than gbrt-rpo-subsampled.

In friedman1-ind, where there is no correlation between the outputs, the best
strategy is single target which makes independent models for each output.  From
a conceptual and statistical point of view, there is no difference between
artificial-gbrt and st-gbrt. The gbmort algorithm, which is optimal when all
outputs are correlated, is here significantly worse than all other methods. The
two boosting methods with output subsampling (gbrt-rpo-subsampled and
gbrt-relabel-rpo-subsampled method), which can adapt themselves to the absence
of correlation between the outputs, perform better than gbmort, but they are
significantly worse than st-gbrt. For these two algorithms, we note that not
relabelling the leaves (gbrt-rpo-subsampled) leads to superior performance
than relabelling them (gbrt-relabel-rpo-subsampled). Since in
friedman1-ind the outputs have disjoint feature support, the test nodes of a
decision tree fitted on one output will partition the samples using these
features. Thus, it is not suprising that relabelling the tree leaves actually
deteriorates performance.

In the previous experiment, all the outputs were dependent of the inputs.
However in multi-output tasks with very high number of outputs, it is likely
that some of them have few or no links with the inputs, i.e., are pure
noise. Let us repeat the previous experiments with the main difference that we
add to the original 16 outputs 16 purely noisy outputs obtained through random
permutations of the original outputs. We show the results of optimizing each
algorithm in Table~\ref{table:raw-results-noisy-out} and the associated
p-values in Table~\ref{tab:tstudent-friedman1-noisy-out1}. We report the
macro-$r^2$ score computed either on all outputs (macro-$r^2$) including the noisy
outputs or only on the 16 original outputs (half-macro-$r^2$).  P-values were
computed between each pair of algorithms using Student's $t$-test on the macro
$r^2$ score computed on all outputs.

We observe that the gbrt-rpo-subsampled algorithm has the best performance on
friedman1-chain and friedman1-group and is the second best on the
friedman1-ind, below st-gbrt.  Interestingly on friedman1-chain and
friedman1-group, this algorithm is significantly better than all the others,
including gbmort. Since this latter method tries to fit all outputs
simultaneously, it is the most disadvantaged by the introduction of the noisy
outputs.

% python analyse.py -d friedman1_chain_v2_n_out_16_noise_1.0_noisy-out friedman1_mo_16_16_1_noisy-out friedman1_mo_16_4_4_noisy-out friedman1_mo_v3_16_1_16_noisy-out --to-latex

\begin{table}
\caption{Friedman datasets with noisy outputs.}\small
\label{table:raw-results-noisy-out}
\centering
\begin{tabular}{lrl}
\toprule
friedman1-chain  &  half-macro-$r^2$ &           macro-$r^2$ \\
\midrule
st-gbrt                     &       0.611 (4) &  $0.265 \pm 0.006$ (4) \\
gbmort                      &       0.617 (3) &  $0.291 \pm 0.012$ (3) \\
gbrt-relabel-rpo-subsampled &       0.628 (2) &  $0.292 \pm 0.006$ (2) \\
gbrt-rpo-subsampled         &       0.629 (1) &  $0.303 \pm 0.007$ (1) \\

\midrule
Friedman1-group &  half-macro-$r^2$ &           macro-$r^2$ \\
\midrule
st-gbrt                     &       0.840 (3) &  $0.364 \pm 0.007$ (4) \\
gbmort                      &       0.833 (4) &  $0.394 \pm 0.004$ (3) \\
gbrt-relabel-rpo-subsampled &       0.855 (2) &  $0.395 \pm 0.005$ (2) \\
gbrt-rpo-subsampled         &       0.862 (1) &  $0.414 \pm 0.006$ (1) \\
\midrule
Friedman1-ind &  half-macro-$r^2$ &           macro-$r^2$ \\
\midrule
st-gbrt                     &       0.806 (1) &  $0.3536 \pm 0.0015$ (1) \\
gbmort                      &       0.486 (4) &   $0.1850 \pm 0.0081$ (4) \\
gbrt-relabel-rpo-subsampled &       0.570 (3) &  $0.2049 \pm 0.0033$ (3) \\
gbrt-rpo-subsampled         &       0.739 (2) &  $0.3033 \pm 0.0021$ (2) \\

\bottomrule
\end{tabular}
\end{table}

\begin{table}
  \caption{P-values given by Student's paired $t$-test on the
    synthetic datasets with additional noisy outputs.  We highlight p-values
    inferior to $\alpha=0.05$ in bold. Note that the sign $>$
    (resp. $<$) indicates that the estimator in the row has better
    (resp.lower) score than the column estimator.}
\label{tab:tstudent-friedman1-noisy-out1}
\centering
% \begin{small}
\begin{tabular}{lllll}
% \toprule
{} & \rotatebox{90}{st-gbrt} & \rotatebox{90}{gbmort} & \rotatebox{90}{gbrt-relabel-rpo-subsampled} & \rotatebox{90}{gbrt-rpo-subsampled} \\
\midrule
Dataset friedman1-chain \\
\midrule
st-gbrt                     &                         &     \textbf{0.0009}(<) &                          \textbf{0.0002}(<) &                  \textbf{0.0002}(<) \\
gbmort                      &      \textbf{0.0009}(>) &                        &                                        0.86 &                    \textbf{0.04}(<) \\
gbrt-relabel-rpo-subsampled &      \textbf{0.0002}(>) &                   0.86 &                                             &                    \textbf{0.03}(<) \\
gbrt-rpo-subsampled         &      \textbf{0.0002}(>) &       \textbf{0.04}(>) &                            \textbf{0.03}(>) &                                     \\
\midrule
Dataset friedman1-group \\
\midrule
st-gbrt                     &                         &     \textbf{0.0002}(<) &                          \textbf{0.0006}(<) &                  \textbf{0.0003}(<) \\
gbmort                      &      \textbf{0.0002}(>) &                        &                                        0.74 &                   \textbf{0.008}(<) \\
gbrt-relabel-rpo-subsampled &      \textbf{0.0006}(>) &                   0.74 &                                             &                   \textbf{0.002}(<) \\
gbrt-rpo-subsampled         &      \textbf{0.0003}(>) &      \textbf{0.008}(>) &                           \textbf{0.002}(>) &                                     \\
\midrule
Dataset friedman1-ind \\
\midrule
st-gbrt                     &                         &      \textbf{1e-06}(>) &                           \textbf{1e-07}(>) &                   \textbf{1e-06}(>) \\
gbmort                      &       \textbf{1e-06}(<) &                        &                            \textbf{0.02}(<) &                   \textbf{6e-06}(<) \\
gbrt-relabel-rpo-subsampled &       \textbf{1e-07}(<) &       \textbf{0.02}(>) &                                             &                   \textbf{2e-06}(<) \\
gbrt-rpo-subsampled         &       \textbf{1e-06}(<) &      \textbf{6e-06}(>) &                           \textbf{2e-06}(>) &                                     \\
\bottomrule
\end{tabular}
% \end{small}
\end{table}

\subsection{Effect of random projection}
\label{sec:gb-exp-rp-effect}

With the gradient boosting and random projection of the output space approaches
(Algorithms~\ref{algo:gbrt-rp} and~\ref{algo:gbrt-rp-relabel}), we have
considered until now only sub-sampling a single output at each iteration as
random projection scheme. In Section~\ref{sec:exp-projections}, we show
empirically the effect of other random projection schemes such as Gaussian
random projection. In Section~\ref{sec:exp-rp-size}, we study the effect of
increasing the number of projections in the gradient boosting algorithm with
random projection of the output space and relabelling
(parameter $q$ of Algorithm~~\ref{algo:gbrt-rp-relabel}). We also show empirically the link
between Algorithm~~\ref{algo:gbrt-rp-relabel} and gradient boosting with
multi-output regression tree (Algorithm~\ref{algo:gb-mo}).

\subsubsection{Choice of the random projection scheme}
\label{sec:exp-projections}

Beside random output sub-sampling, we can combine the multi-output gradient boosting strategies
(Algorithms~\ref{algo:gbrt-rp} and~\ref{algo:gbrt-rp-relabel}) with other random
projection schemes. A key difference between
random output sub-sampling and random projections such as Gaussian and (sparse)
Rademacher projections is that the latter combines together several outputs.

\begin{figure}
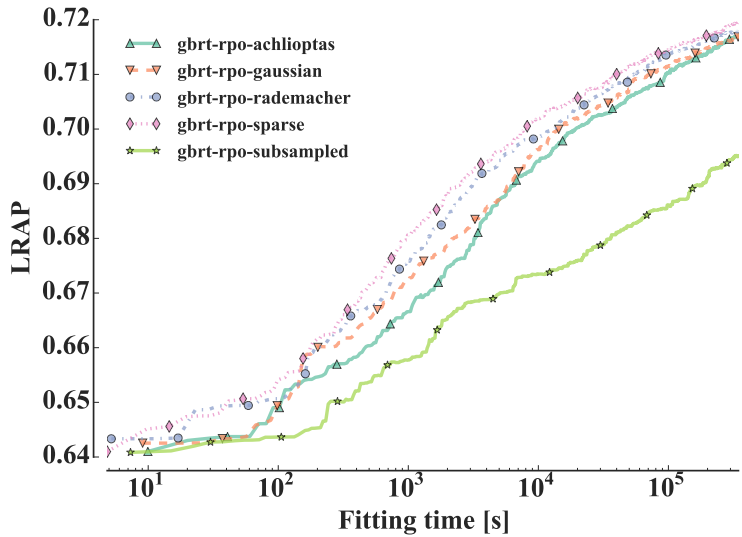

\centering
\includegraphics[width=0.7\textwidth]{{{projection-norelabel_mediamill_log_time_lrap}}}
\caption{
On the mediamill dataset, Gaussian, Achlioptas and sparse random projections with
gbrt-rpo (Algorithm \ref{algo:gbrt-rp}) show 10 times faster convergence in
terms of LRAP score, than sub-sampling one output variable at each iteration.
($k=p$, stumps, $\mu=0.1$, logistic loss)}
\label{fig:mediamill-projections}
\end{figure}

\begin{figure}
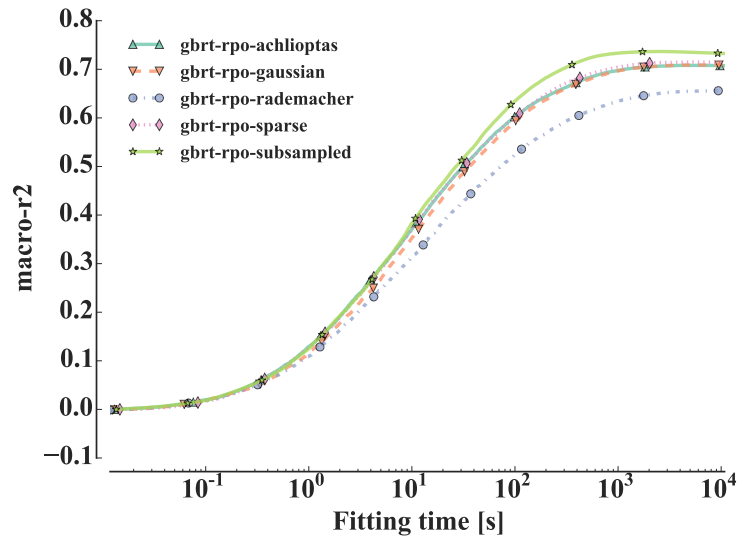

\centering
\includegraphics[width=0.7\textwidth]{{{projection-norelabel_friedman1_mo_16_16_1_log_time_macro-r2}}}
\caption{
On the friedman1-ind dataset where there is no output correlation, gbrt-rpo
(Algorithm~\ref{algo:gbrt-rp}) with one random subsampled output leads to a higher
macro-$r^2$ score than using Gaussian, Achlioptas or sparse random projections.
($k=p$, stumps, $\mu=0.1$, square loss)
\label{fig:friedman1-ind-projections}}
\end{figure}

We show in Figures~\ref{fig:mediamill-projections}, \ref{fig:delicious-projections} and
 \ref{fig:friedman1-ind-projections} the LRAP or macro-$r^2$ score
convergence of gradient boosting with randomly projected outputs (gbrt-rpo,
Algorithm~\ref{algo:gbrt-rp}) respectively on the mediamill, delicious, and Friedman1-ind
datasets with different random projection schemes.

The impact of the random projection scheme on convergence speed of gbrt-rpo
(Algorithm~\ref{algo:gbrt-rp}) is very problem dependent. On the mediamill
dataset, Gaussian, Achlioptas, or sparse random projections all improve
convergence speed by a factor of 10 (see
Figure~\ref{fig:mediamill-projections}) compared to subsampling randomly only
one output. On the delicious (Figure~\ref{fig:delicious-projections}) and
friedman1-ind (Figure~\ref{fig:friedman1-ind-projections}), this is the
opposite: subsampling leads to faster convergence than all other projections
schemes.  Note that we have the same behavior if one relabels the tree
structure grown at each iteration as in Algorithm~\ref{algo:gbrt-rp-relabel}
(results not shown).

Dense random projections, such as Gaussian random projections, force the weak
model to consider several outputs jointly and it should thus only improve when
outputs are somewhat correlated (which seems to be the case on mediamill). When
all of the outputs are independent or the correlation is less strong, as in
friedman1-ind or delicious, this has a detrimental effect. In this situation,
sub-sampling only one output at each iteration leads to the best performance.

% python analyse_plot_time.py --relabel False --no-baseline --dir projection-norelabel -d mediamill friedman1_mo_16_16_1 delicious

\begin{figure}
\centering
\includegraphics[width=0.7\textwidth]{{{projection-norelabel_delicious_log_time_lrap}}}
\caption{
On the delicious dataset, Gaussian, Achlioptas and sparse random projections with
gbrt-rpo (Algorithm~\ref{algo:gbrt-rp}) show 10 times faster convergence in
terms of LRAP score, than sub-sampling one output variable at each iteration.
($k=p$, stumps, $\mu=0.1$, logistic loss)}
\label{fig:delicious-projections}
\end{figure}

\subsubsection{Effect of the size of the projected space}
\label{sec:exp-rp-size}

The multi-output gradient boosting strategy combining random projections and
tree relabelling (Algorithm~\ref{algo:gbrt-rp-relabel}) can use  more than one
random projection ($q\geq 1$) by using multi-output trees as base learners. In
this section, we study the effect of the size of the projected space $q$ in
Algorithm~\ref{algo:gbrt-rp-relabel}. This approach corresponds to the one
developed in~\cite{joly2014random} for random forest transposed to gradient boosting.

\begin{figure}
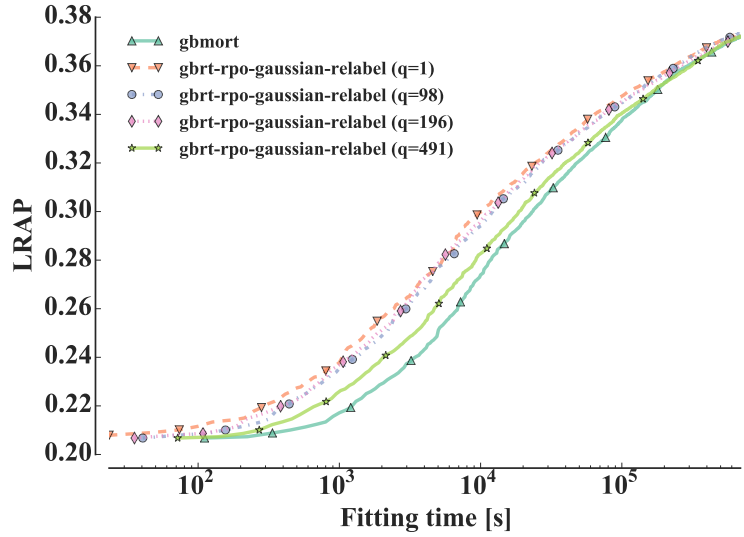
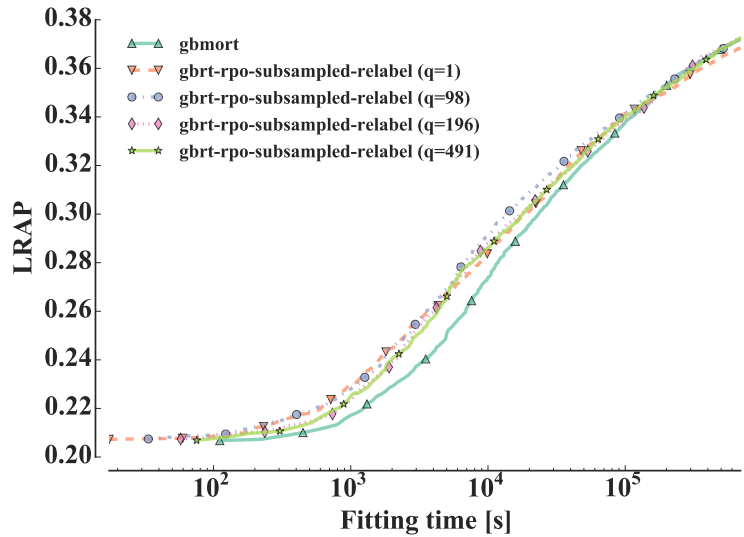

\centering
\subfloat[]{\label{subfig:delicious-time-gaussian}\includegraphics[width=0.7\textwidth]{{{n_comp_gaussian_delicious_log_time_lrap_lr=0.1}}}} \\
\subfloat[]{\label{subfig:delicious-time-subsample}\includegraphics[width=0.7\textwidth]{{{n_comp_subs_delicious_log_time_lrap_lr=0.1}}}}
\caption{On the delicious dataset, LRAP score as a function of the boosting
ensemble fitting time for  gbrt-rpo-gaussian-relabel and
gbrt-rpo-subsampled-relabel with different number of projections $q$.
($k=p$, stumps, $\mu=0.1$, logistic loss)}
\label{fig:delicious-time-q}
\end{figure}

Figure~\ref{fig:delicious-time-q} shows the LRAP score as a function of the
fitting time for gbmort (Algorithm~\ref{algo:gb-mo}) and gbrt-relabel-rpo
(Algorithm~\ref{algo:gbrt-rp-relabel}) with either Gaussian random projection
(see Figure \ref{subfig:delicious-time-gaussian}) or output subsampling (see
Figure~\ref{subfig:delicious-time-subsample}) for a number of projections $q
\in \left\{1, 98, 196, 491\right\}$ on the delicious dataset. In
Figure~\ref{subfig:delicious-time-gaussian} and
Figure~\ref{subfig:delicious-time-subsample}, one Gaussian random projection or
one sub-sampled output has faster convergence than their counterparts with a
higher number of projections $q$ and gbmort at fixed computational budget. Note
that when the number of projections $q$ increases, gradient boosting with
random projection of the output space and relabelling becomes similar to gbmort.

% python analyse_plot_time_n_components.py -t subsampled -d delicious
% python analyse_plot_time_n_components.py -t gaussian -d delicious

\begin{figure}
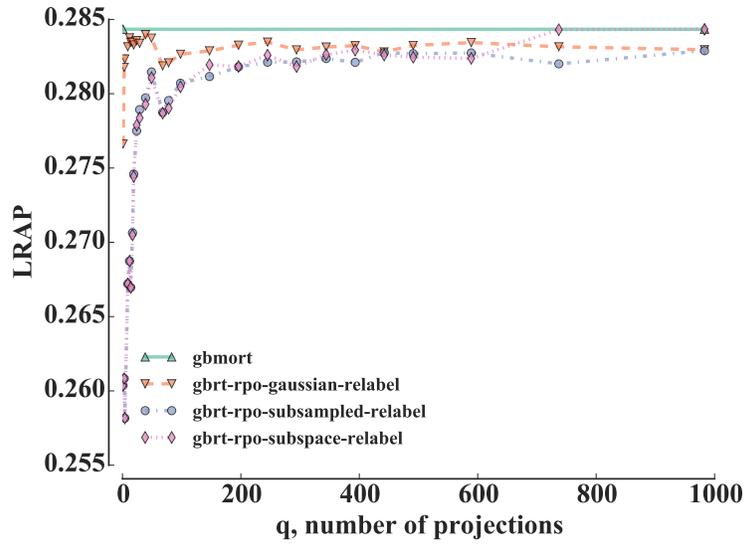
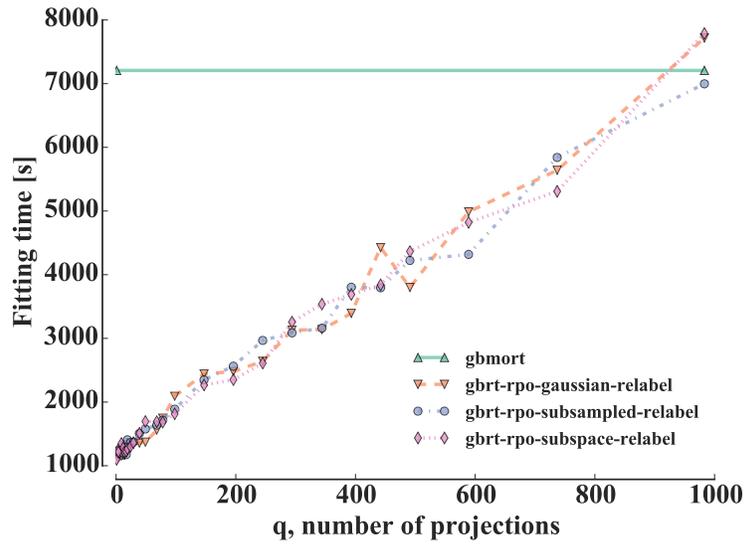

\centering
\subfloat[]{\label{subfig:rp-delicious-lrap}\includegraphics[width=0.7\textwidth]{{{n_comp_delicious_lrap_loss=logistic-brent}}}} \\
\subfloat[]{\label{subfig:rp-delicious-time}\includegraphics[width=0.7\textwidth]{{{n_comp_delicious_chrono-fit_loss=logistic-brent}}}}
\caption{On delicious, increasing the number of random projections $q$ allows
         to reach the same LRAP score as gbmort at a significantly
         reduced computational cost.
         ($k=p$, stumps, $\mu=0.1$, $M=100$, logistic loss)}
\label{fig:increasing-rp}
\end{figure}

Instead of fixing the computational budget as a function of the training time,
we now set the computational budget to 100 boosting steps. On the delicious
dataset, gbrt-relabel-rpo (Algorithm~\ref{algo:gbrt-rp-relabel}) with Gaussian
random projection yields approximately the same performance as gbmort with
$q\geq20$ random projections as shown in Figure~\ref{subfig:rp-delicious-lrap}
and reduces computing times by \emph{a factor of 7} at $q=20$ projections (see
Figure~\ref{subfig:rp-delicious-time}).

% python analyse_plot_n_components.py -d delicious

These experiments show that gradient boosting with random projection and
relabelling (gbrt-relabel-rpo, Algorithm~\ref{algo:gbrt-rp-relabel}) is indeed an
approximation of gradient boosting with multi-output trees (gbmort,
Algorithm~\ref{algo:gb-mo}). The number of random projections $q$ influences
simultaneously the bias-variance tradeoff and the convergence speed of
Algorithm~\ref{algo:gbrt-rp-relabel}.

\subsection{Systematic analysis over real world datasets}
\label{sec:systematic-analysis}

We perform a systematic analysis over real world multi-label classification and multi-output
regression datasets. For this study, we evaluate the proposed algorithms:
gradient boosting of multi-output regression trees (gbmort,
Algorithm~\ref{algo:gb-mo}), gradient boosting with random projection of the
output space (gbrt-rpo, Algorithm~\ref{algo:gbrt-rp}), and gradient boosting
with random projection of the output space  and relabelling (gbrt-relabel-rpo,
Algorithm~\ref{algo:gbrt-rp-relabel}). For the two latter algorithms, we
consider two random projection schemes: (i) Gaussian random projection, a dense
random projection, and (ii) random output sub-sampling, a sparse random
projection. They will be compared to three common and well established
tree-based multi-output algorithms: (i) binary relevance / single target of
gradient boosting regression tree (br-gbrt / st-gbrt), (ii) multi-output random
forest (mo-rf) and (iii) binary relevance / single target of random forest
models (br-rf / st-rf).

We will compare all methods on multi-label tasks in
Section~\ref{subsec:gb-multilabel-exp} and on multi-output regression
tasks in Section~\ref{subsec:gb-mo-reg-exp}. Following the
recommendations in \citep{demvsar2006statistical}, we use the Friedman
test and its associated Nemenyi post-hoc test to assess the
statistical significance of the observed differences. Pairwise
comparisons are also carried out using the Wilcoxon signed ranked
test.

\subsubsection{Multi-label datasets}
\label{subsec:gb-multilabel-exp}

\begin{figure}
\caption{Critical difference diagram between algorithms on the multi-label datasets.}
\label{fig:cd-multilabel}
\centering
\includegraphics[width=\textwidth]{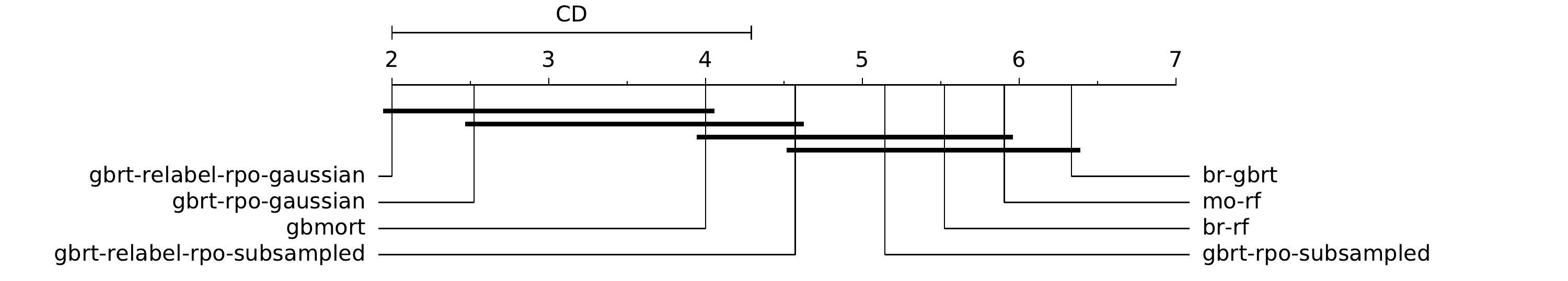}
\end{figure}

The critical distance diagram of Figure~\ref{fig:cd-multilabel} gives
the ranks of the algorithms over the 21 multi-label datasets and has
an associated Friedman test p-value of $1.36 \times 10^{-10}$ with a
critical distance of $2.29$ given by the Nemenyi post-hoc test
($\alpha=0.05$). Thus, we can reject the null hypothesis that all
methods are equivalent.  Table~\ref{tab:wilcoxon-multilabel} gives the
outcome of the pairwise Wilcoxon signed ranked tests. Detailed scores
are provided in Appendix~\ref{appendix:overall-mulit-label}.

The best average performer is gbrt-relabel-rpo-gaussian which is significantly
better according to the Nemenyi post-hoc test than all methods except
gbrt-rpo-gaussian and gbmort.

Gradient boosting with the Gaussian random projection has a significantly
better average rank than the random output sub-sampling projection. Relabelling
tree leaves allows to have better performance on the 21 multi-label dataset.
Indeed, both gbrt-relabel-rpo-gaussian and gbrt-relabel-rpo-subsampled are
better ranked and significantly better than their counterparts without
relabelling (gbrt-rpo-gaussian and gbrt-rpo-subsampled).

%{\color{red} (PG: \`a discuter et r\'e\'ecrire)}
Among all compared methods, br-gbrt has the worst rank and it is significantly
worse than all gbrt variants according to the Wilcoxon signed rank
test. This might be actually a consequence of the constant budget in time that
was allocated to all methods (see Section~\ref{sec:experimental-protocol}). All
methods were given the same budget in time but, given the very slow convergence
rate of br-gbrt, this budget may not allow to grow enough trees per output with
this method to reach competitive performance.

We notice also that both random forests based methods (mo-rf and br-rf) are
less good than all gbrt variants, most of the time significantly, except for
br-gbrt. It has to be noted however that no hyper-parameter was tuned for the
random forests. Such tuning could slightly change our conclusions, although
random forests often work well with default setting.

\begin{table}
% \begin{tiny}
\caption{P-values given by the Wilcoxon signed rank test on the multi-label datasets.
We bold p-values below $\alpha=0.05$. Note that the sign $>$ (resp. $<$) indicates that
the row estimator has superior (resp. inferior) LRAP score than the column estimator.}
\label{tab:wilcoxon-multilabel}
\centering
\setlength{\tabcolsep}{1pt}
% \rotatebox{90}{
\begin{small}
\begin{tabular}{lllllllll}
% \toprule
{} & \rotatebox{90}{br-gbrt} & \rotatebox{90}{br-rf} & \rotatebox{90}{gbmort} & \rotatebox{90}{gbrt-relabel-rpo-gaussian} & \rotatebox{90}{gbrt-relabel-rpo-subsampled} & \rotatebox{90}{gbrt-rpo-gaussian} & \rotatebox{90}{gbrt-rpo-subsampled} & \rotatebox{90}{mo-rf} \\
% \midrule
br-gbrt                     &                         &                  0.79 &      \textbf{0.001}(<) &                         \textbf{6e-05}(<) &                           \textbf{0.001}(<) &                \textbf{0.0001}(<) &                   \textbf{0.003}(<) &                  0.54 \\
br-rf                       &                    0.79 &                       &       \textbf{0.01}(<) &                         \textbf{0.002}(<) &                            \textbf{0.02}(<) &                 \textbf{0.005}(<) &                                0.07 &                  0.29 \\
gbmort                      &       \textbf{0.001}(>) &      \textbf{0.01}(>) &                        &                         \textbf{0.001}(<) &                                        0.36 &                  \textbf{0.04}(<) &                    \textbf{0.03}(>) &      \textbf{0.02}(>) \\
gbrt-relabel-rpo-gaussian   &       \textbf{6e-05}(>) &     \textbf{0.002}(>) &      \textbf{0.001}(>) &                                           &                          \textbf{0.0002}(>) &                 \textbf{0.005}(>) &                   \textbf{7e-05}(>) &    \textbf{0.0007}(>) \\
gbrt-relabel-rpo-subsampled &       \textbf{0.001}(>) &      \textbf{0.02}(>) &                   0.36 &                        \textbf{0.0002}(<) &                                             &                \textbf{0.0002}(<) &                    \textbf{0.02}(>) &     \textbf{0.008}(>) \\
gbrt-rpo-gaussian           &      \textbf{0.0001}(>) &     \textbf{0.005}(>) &       \textbf{0.04}(>) &                         \textbf{0.005}(<) &                          \textbf{0.0002}(>) &                                   &                   \textbf{7e-05}(>) &     \textbf{0.002}(>) \\
gbrt-rpo-subsampled         &       \textbf{0.003}(>) &                  0.07 &       \textbf{0.03}(<) &                         \textbf{7e-05}(<) &                            \textbf{0.02}(<) &                 \textbf{7e-05}(<) &                                     &      \textbf{0.04}(>) \\
mo-rf                       &                    0.54 &                  0.29 &       \textbf{0.02}(<) &                        \textbf{0.0007}(<) &                           \textbf{0.008}(<) &                 \textbf{0.002}(<) &                    \textbf{0.04}(<) &                       \\
\end{tabular}
\end{small}
% \end{tiny}
% }
\end{table}

\subsubsection{Multi-output regression datasets}
\label{subsec:gb-mo-reg-exp}

\begin{figure}
\caption{Critical difference diagram between algorithm on the multi-output regression
datasets.}
\label{fig:cd-regression}
\centering
\includegraphics[width=\textwidth]{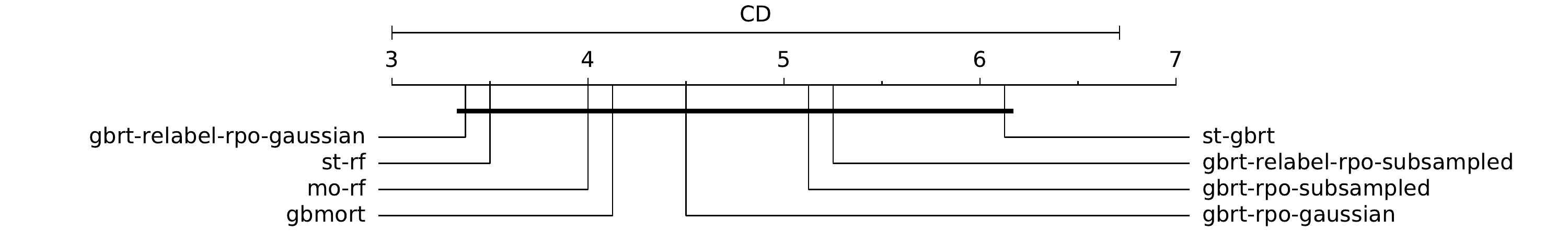}
\end{figure}

The critical distance diagram of Figure~\ref{fig:cd-regression} gives the rank
of each estimator over the 8 multi-output regression datasets. The
associated Friedman test has a p-value of $0.3$. Given the outcome of the test,
we can therefore not reject the null hypothesis that the estimator performances
can not be distinguished. Table~\ref{tab:wilcoxon-regression} gives the outcomes
of the pairwise Wilcoxon signed ranked tests. They confirm the fact that all
methods are very close to each other as only two comparisons show a p-value
lower than 0.05 (st-rf is better than st-gbrt and gbrt-rpo-subsampled). This
lack of statistical power is probably partly due here to the smaller number of
datasets included in the comparison (8 problems versus 21 problems in
classification and 8 methods tested over 8 problems).  Please, see
Appendix~\ref{appendix:overall-multi-output-regression} for the detailed scores.

If we ignore statistical tests, as with multi-label tasks,
gbrt-relabel-rpo-gaussian has the best average rank and st-gbrt the worst
average rank. This time however, gbrt-relabel-rpo-gaussian is followed by the
random forest based algorithms (st-rf and mo-rf) and gbmort. Given the lack of
statistical significance, this ranking should however be intrepreted cautiously.

\begin{table}
\caption{P-value given by the Wilcoxon signed rank test on multi-output
regression datasets. We bold p-values below $\alpha=0.05$. Note that the sign $>$ (resp. $<$) indicates that
the row estimator has superior (resp. inferior) macro-$r^2$ score than the column estimator.}
\label{tab:wilcoxon-regression}
\centering
% \setlength{\tabcolsep}{2pt}
% \begin{footnotesize}
\begin{tabular}{lllllllll}
{} & \rotatebox{90}{st-gbrt} & \rotatebox{90}{st-rf} & \rotatebox{90}{gbmort} & \rotatebox{90}{gbrt-relabel-rpo-gaussian} & \rotatebox{90}{gbrt-relabel-rpo-subsampled} & \rotatebox{90}{gbrt-rpo-gaussian} & \rotatebox{90}{gbrt-rpo-subsampled} & \rotatebox{90}{mo-rf} \\
st-gbrt                     &                         &      \textbf{0.02}(<) &                   0.26 &                                      0.58 &                                        0.78 &                              0.67 &                                0.89 &                  0.16 \\
st-rf                       &        \textbf{0.02}(>) &                       &                   0.33 &                                      0.67 &                                        0.09 &                              0.09 &                    \textbf{0.04}(>) &                  0.58 \\
gbmort                      &                    0.26 &                  0.33 &                        &                                       0.4 &                                        0.16 &                              0.67 &                                 0.4 &                  0.48 \\
gbrt-relabel-rpo-gaussian   &                    0.58 &                  0.67 &                    0.4 &                                           &                                        0.16 &                               0.4 &                                0.48 &                  0.58 \\
gbrt-relabel-rpo-subsampled &                    0.78 &                  0.09 &                   0.16 &                                      0.16 &                                             &                              0.16 &                                   1 &                  0.07 \\
gbrt-rpo-gaussian           &                    0.67 &                  0.09 &                   0.67 &                                       0.4 &                                        0.16 &                                   &                                0.48 &                  0.26 \\
gbrt-rpo-subsampled         &                    0.89 &      \textbf{0.04}(<) &                    0.4 &                                      0.48 &                                           1 &                              0.48 &                                     &                   0.4 \\
mo-rf                       &                    0.16 &                  0.58 &                   0.48 &                                      0.58 &                                        0.07 &                              0.26 &                                 0.4 &                       \\
\end{tabular}
% \end{footnotesize}
\end{table}

\clearpage

\section{Related works}
\label{sec:rw}

Binary relevance~\cite{tsoumakas2009mining} / single
target~\cite{spyromitros2012multi} is often suboptimal as it does not exploit
potential correlations that might exist between the outputs. For this reason,
several approaches have been proposed in the literature that improve over binary
relevance by exploiting output dependencies. Besides the multi-output tree-based
methods discussed in Sections~\ref{subsec:mo-dt} and \ref{subsec:gb-extend-mo},
with which we compared ourself empirically in Section~\ref{sec:experiments}, we
review in this section several other related algorithms relying either on random
projection of the output space, explicit consideration of output dependencies,
or on the sharing of models learnt for one output with the others. We refer the
interested reader to
\cite{zhang2014review,tsoumakas2009mining,gibaja2014multi,borchani2015survey,spyromitros2012multi}
for more complete reviews of multilabel/multi-output methods, and to
\cite{madjarov2012extensive} for an empirical comparison of these methods.

Given a specific multi-label or multi-output regression loss function, one can
sometimes exploit its mathematical properties to derive a specific Adaboost
algorithm: minimizing at each boosting step the selected loss function by
choosing carefully the sample weights prior training a weak estimator. Each loss
minimizable in such way ending with its own dedicated training algorithm:
Adaboost.MH \cite{schapire1999improved} for the Hamming loss, Adaboost.MR for
the Ranking loss \cite{schapire1999improved}, Adaboost.LC for the covering loss
\cite{amit2007boosting}, AdaBoostSeq \cite{kajdanowicz2013boosting} for the
sequence loss and Adaboost.MRT \cite{kummer2014adaboost} for the $\ell_1$ loss.
By contrast as \cite{friedman2001greedy}, we have instead defined boosting
algorithms applicable to the whole family of differentiable loss functions
through a gradient descent inspired approach. Note that prior multi-output
gradient boosting approach has also focused on either the $\ell_2$ loss
\cite{geurts2007gradient} or the covariance discrepancy loss
\cite{miller2016finding}.

Random projections have been exploited by several authors to solve
multi-output supervised learning
problems. \citet{hsu2009multi,kapoor2012multilabel} address sparse
multi-label problems using compressed sensing techniques. The original
output label vector is projected into a low dimensional space and then
a (linear) regression model is fit on each projected output in the
reduced space. At prediction time, the predictions made by all
estimators are then concatenated and decoded back into the original
space. This latter step requires to solve an inverse problem
exploiting the sparsity of the original label
vector. \citet{tsoumakas2014multi} also train single-output models
(gradient boosted trees) independently on random linear projections of
the output vector. Unlike \cite{hsu2009multi} however, the number of
generated random projections is actually larger than the number of
original outputs, since this is expected to improve accuracy by adding
an ensemble effect, and decoding involves solving an overdetermined
linear system, which can be done without requiring any sparsity
assumption. In our own previous work \cite{joly2014random}, we
investigated random output projections to fasten and improve the
accuracy of multi-output random forests. Each tree of the forest is a
multi-output tree which is fitted on a low-dimensional random
projection of the original outputs, with the projection renewed for
each tree independently. Tree leaves of the resulting forest are then
relabeled in the original output space (as in Algorithm 4). With
respect to
\cite{hsu2009multi,kapoor2012multilabel,tsoumakas2014multi},
relabelling the leaves avoids having to solve an inverse problem at
prediction time, which improves computing times and also reduces the
risk of introducing errors at decoding.

The previous approaches exploit the output correlation only
implicitely in that the individual models are trained to fit random
linear combinations incorporating each several outputs. Individual
models are however trained independently of each other. In our
approach, we take more explicitely output dependencies into account by
iteratively fitting linearly the current random projection estimator
to (the current residuals of) all outputs at each iteration. In this
sense, our work is related to several multi-label or multi-target
methods that try to more explicitely exploit output dependencies or to
convey information from one output to the others. For example, in the estimator chain
approach~\cite{read2011classifier,dembczynski2010bayes,spyromitros2012multi},
the conditional probability $P_{\mathcal{Y}|\mathcal{X}}(y|x)$ is
modeled using the chain rule:
\begin{equation}
 P_{\mathcal{Y}|\mathcal{X}}(y|x) = P_{\mathcal{Y}_1|\mathcal{X}}(y_1|x) \prod_{j=2}^d
 P_{\mathcal{Y}_j|\mathcal{X},\mathcal{Y}_1,\ldots,\mathcal{Y}_{j-1}}(y_j|x,y_1,\ldots,y_{j-1}).
\label{eq:chain-rule-cl-chain}
\end{equation}
The factors of this decomposition are learned in sequence, each one
using the predictions of the previously learnt models as auxiliary
inputs. Like our approach, these methods are thus also able to benefit
from output dependencies to obtain simpler and/or better
estimators. Several works have extended the estimator chain idea by
exploiting other (data induced) factorization of the conditional
output distribution, derived for example from a bayesian network
structure \cite{zhang2010multi} or using the composition
property~\cite{gasse2015optimality}.

Two works at least are based on the idea of sharing models trained for
one output with the others (in the context of multi-label
classification). \citet{huang2012multi} propose to generate an
ensemble through a sequential approach: at each step, one model is
fitted for each output independently without taking into account the
others, and then an optimal linear combination of these models is
fitted to each output so as to minimize the selected loss. The process
is costly as one has to fit several linear models at each
iteration. Convergence speed of this method with respect to the number
of base models is also expected to be slower than ours, because
information between outputs is shared less frequently (every $d$ base
models, with $d$ the number of labels). \citet{yan2007model} propose a
two-step approach. First, a pool of models is generated by training a
single individual classifier for each label independently (using
feature randomization and bootstrapping). A boosted ensemble is then
fit for each label independently using base models selected from the
pool obtained during the first step (each one potentially trained for
a label different from the targeted one). This approach is less
adaptive than others since the pool of base models is fixed, for
efficiency reasons.

Finally, more complementary to our work, \citet{SiZKMDH17} recently
proposed an efficient implementation of a multi-label gradient boosted
decision tree method (very similar to the gbmort method studied here)
for handling very high dimensional and sparse label vectors. While
very effective, this implementation aims at improving prediction time
and model size in the presence of sparse outputs and does not change
the representation bias with respect to gbmort. It should thus suffer
from the same limitations as this method in terms of predictive
performance.

\section{Conclusions}
\label{sec:conclusions}

In this paper, we have first formally extended the gradient boosting
algorithm to multi-output tasks leading to the ``multi-output gradient boosting
algorithm'' (gbmort). It sequentially minimizes a multi-output loss using
multi-output weak models considering that all outputs are correlated. By
contrast, binary relevance / single target of gradient boosting models fits one
gradient boosting model per output considering that all outputs are
independent. However in practice, we do not expect to have either all outputs
independent or all outputs dependent. So, we propose a more flexible approach
which adapts automatically to the output correlation structure called
``gradient boosting with random projection of the output space'' (gbrt-rpo). At
each boosting step, it fits a single weak model on a random projection of the
output space and optimize a multiplicative weight separately for each output.
We have also proposed a variant of this algorithm (gbrt-relabel-rpo) only valid
with decision trees as weak models: it fits a decision tree on the randomly
projected space and then it relabels tree leaves with predictions in the
original (residual) output space. The combination of the gradient boosting
algorithm and the random projection of the output space yields faster
convergence by exploiting existing correlations between the outputs and by
reducing the dimensionality of the output space. It also provides new
bias-variance-convergence trade-off potentially allowing to improve
performance.

We have evaluated in depth these new algorithms on several artificial and real
datasets. Experiments on artificial problems highlighted that gb-rpo with output
subsampling offers an interesting tradeoff between single target and
multi-output gradient boosting. Because of its capacity to automatically adapt
to the output space structure, it outperforms both methods in terms of
convergence speed and accuracy when outputs are dependent and it is superior to
gbmort (but not st-rt) when outputs are fully independent. On the 29 real
datasets, gbrt-relabel-rpo with the denser Gaussian projections turns out to be
the best overall approach on both multi-label classification and multi-output
regression problems, although all methods are statistically undistinguisable on
the regression tasks. Our experiments also show that gradient boosting based
methods are competitive with random forests based methods. Given that
multi-output random forests were shown to be competitive with several other
multi-label approaches in \cite{madjarov2012extensive}, we are confident that
our solutions will be globally competitive as well, although a broader empirical
comparison should be conducted as future work. One drawback of gradient boosting
with respect to random forests however is that its performance is more sensitive
to its hyper-parameters that thus require careful tuning. Although not discussed
in this paper, besides predictive performance, gbrt-rpo (without relabeling) has
also the advantage of reducing model size with respect to mo-rf (multi-output
random forests) and gbmort, in particular in the presence of many outputs.
Indeed, in mo-rf and gbmort, one needs to store a vector of the size of the
number of outputs per leaf node. In gbrt-rpo, one needs to store only one real
number (a prediction for the projection) per leaf node and a vector of the size
of the number of outputs per tree ($\rho_m$). At fixed number of trees and fixed
tree complexity, this could lead to a strong reduction of the model memory
requirement when the number of labels is large. Note that the approach proposed
in~\cite{joly2014random} does not solve this issue because of leaf node
relabeling. This could be addressed by deactivating leaf relabeling and
inverting the projection at prediction time to obtain a prediction in the
original output space, as done for example in
\cite{hsu2009multi,kapoor2012multilabel,tsoumakas2014multi}. However, this would
be at the expense of computing times at prediction time and of accuracy because
of the potential introduction of errors at the decoding stage. Finally, while we
restricted our experiments here to tree-based weak learners,
Algorithms~\ref{algo:gb-mo} and \ref{algo:gbrt-rp} are generic and could exploit
respectively any multiple output and any single output regression method. As
future work, we believe that it would interesting to evaluate them with other
weak learners.

\section{Acknowledgements}

% \begin{acknowledgements}
Part of this research has been done while A. Joly was a research
fellow of the FNRS, Belgium.  This work is partially supported by the
IUAP DYSCO, initiated by the Belgian State, Science Policy Office.
% \end{acknowledgements}

\appendix

\section{Convergence when $M \rightarrow \infty$} %LW: Large M rather than small m, I guess
\label{sec:convergence-proof}

Similarly to~\cite{geurts2007gradient}, we can prove the convergence of the
training-set loss of the gradient boosting with multi-output models
(Algorithm~\ref{algo:gb-mo}), and gradient boosting on randomly projected spaces
with (Algorithm~\ref{algo:gbrt-rp}) or without relabelling
(Algorithm~\ref{algo:gbrt-rp-relabel}).

Since the loss function is lower-bounded by $0$, we merely need to show that the loss $\ell$ is
non-increasing on the training set at each step $m$ of the gradient boosting
algorithm.

For Algorithm~\ref{algo:gb-mo} and Algorithm~\ref{algo:gbrt-rp-relabel}, we
note that
\begin{align}
\sum_{i=1}^n \ell(y^i, f_{m}(x^i))
&= \min_{\rho \in \mathbb{R}^{d}} \sum_{i=1}^n \ell\left(y^i, f_{m-1}(x^i) + \rho\odot g_m(x^i)\right) \nonumber \\
&\leq \sum_{i=1}^n \ell\left(y^i, f_{m-1}(x^i) \right).\label{thm:conv-rho}
\end{align}
and the learning-set loss is hence non increasing with $M$ if we use a learning
rate $\mu=1$. If the loss $\ell(y,y')$ is convex in its second
argument $y'$ (which is the case for those loss-functions that we use in
practice), then this convergence property actually holds for any value
$\mu\in(0;1]$ of the learning rate. Indeed, we have
\begin{eqnarray*}
& & \sum_{i=1}^n \ell\left(y^i, f_{m-1}(x^i) \right)\\
& \geq &(1-\mu) \sum_{i=1}^n \ell\left(y^i, f_{m-1}(x^i) \right) + \mu \sum_{i=1}^n \ell\left(y^i, f_{m-1}(x^i) + \rho_m \odot g_m(x^i)\right)\\
& \geq & \sum_{i=1}^n \ell\left(y^i, f_{m-1}(x^i) + \mu \rho_m \odot g_m(x^i)\right).
\end{eqnarray*}
\noindent given Equation~\ref{thm:conv-rho} and the convexity property.

For Algorithm~\ref{algo:gbrt-rp}, we have a weak estimator $g_m$ fitted on a
single random projection of the output space $\phi_m$ with a multiplying
constant vector $\rho_m \in \mathbb{R}^d$, and we have:
\begin{align}
\sum_{i=1}^n \ell(y^i, f_{m}(x^i))
&=  \min_{\rho \in \mathbb{R}^{d}} \sum_{i=1}^n \ell\left(y^i, f_{m-1}(x^i) + \rho g_m(x^i)\right) \nonumber \\
&\leq \sum_{i=1}^n \ell\left(y^i, f_{m-1}(x^i)\right).
\end{align}
and the error
is also non increasing for Algorithm~\ref{algo:gbrt-rp}, under the same conditions as above.

The previous development shows that Algorithm~\ref{algo:gb-mo},
Algorithm~\ref{algo:gbrt-rp} and Algorithm~\ref{algo:gbrt-rp-relabel} are
converging on the training set for a given loss $\ell$. The binary relevance /
single target of gradient boosting regression trees admits a similar convergence
proof. We expect however  the convergence speed of the binary relevance /
single target to be lower assuming that it fits one weak estimator for each
output in a round robin fashion.

\section{Representation bias of decision tree ensembles}
\label{subsec:gb-rp-discussions}

Random forests and gradient tree boosting build an ensemble of  trees
either independently or sequentially, and thus offer different bias/variance tradeoffs.
The predictions of all these ensembles can be expressed as a weighted combination of
the ground truth outputs of the training set samples. In the present section, we discuss the differences
between single tree models, random forest models and gradient tree boosting
models in terms of the representation biases of the obtained models. We also
highlight the differences between single target models and multi-output tree
models.

\paragraph{Single tree models.}  The prediction of a regression tree learner can
be written as a weighted linear combination of the training samples
$\mathcal{L}=\{(x^i,y^i) \in \mathcal{X} \times \mathcal{Y}\}_{i=1}^n$. We
associate to each training sample $(x^i, y^i)$ a weight function $w^i:
\mathcal{X} \rightarrow \mathbb{R}$ which gives the contribution of a ground
truth $y^i$ to predict an unseen sample $x$. The prediction of a \emph{single output
 tree} $f$ is given by
\begin{equation}
f(x) = \sum_{i=1}^n w^i(x) y^i.
\label{eq:so-dt-pred}
\end{equation}
\noindent The weight function $w^i(x)$ is non zero if both the samples $(x^i,y^i)$
and the unseen sample $x$ reach the same leaf of the tree. If both
$(x^i,y^i)$ and $x$ end up in the same leaf of the  tree,
$w^i(x)$ is equal to the inverse of the number of training samples reaching that leaf.
The weight $w^i(x)$ can thus be rewritten as $k(x^i, x)$ and the function $k(\cdot, \cdot)$ is actually a positive semi-definite
kernel~\cite{geurts2006extremely}.

We can also express multi-output models as a weighted sum of the training
samples. With a \emph{single target regression tree}, we have an
independent weight function $w^i_j$ for each sample $i$ of the training set and
each output $j$ as we fit one model per output. The prediction of this model for output $j$ is
given by:
\begin{equation}
f(x)_j = \sum_{i=1}^n w^i_j(x) y^i_j.
\label{eq:br-dt-pred}
\end{equation}

With a \emph{multi-output regression tree}, the decision tree structure is shared
between all outputs so we have a single weight function $w^i$ for each training
sample $i$:
\begin{equation}
f(x)_j = \sum_{i=1}^n w^i(x) y^i_j.
\label{eq:mo-dt-pred}
\end{equation}

\paragraph{Random forest models.} If we have a \emph{single target
random forest model}, the prediction of the $j$-th output combines the predictions
of the $M$ models of the ensemble in the following way:
\begin{equation}
f(x)_j = \frac{1}{M}\sum_{m=1}^M \sum_{i=1}^n w^{i}_{m,j}(x) y^i_j,
\label{eq:br-rf-pred}
\end{equation}
with one weight function $w^{i}_{m,j}$ per tree, sample and output. We note that
we can combine the weights of the individual trees into a single weight per sample and per output
\begin{equation}
w^i_j(x) = \frac{1}{M} \sum_{m=1}^M w^{i}_{m,j}(x).
\end{equation}
\noindent The prediction of the $j$-th output for an ensemble of independent
models has thus the same form as a single target regression tree
model:
\begin{equation}
f(x)_j = \frac{1}{M}\sum_{m=1}^M \sum_{i=1}^n w^{i}_{m,j}(x) y^i_j = \sum_{i=1}^n w^i_j(x) y^i_j.
\end{equation}

We can repeat the previous development with a \emph{multi-output random forest
model}. The prediction for the $j$-th output of an unseen sample $x$ combines
the predictions of the $M$ trees:
\begin{align}
f(x)_j &= \frac{1}{M}\sum_{m=1}^M \sum_{i=1}^n w^i_m(x) y^i_j = \sum_{i=1}^n w^i(x) y^i_j
\label{eq:mo-rf-pred}
\end{align}
\noindent with
\begin{equation}
w^i(x) = \frac{1}{M} \sum_{m=1}^M w^i_m(x).
\end{equation}
\noindent With this framework, the prediction of an ensemble model has the same
form as the prediction of a single constituting tree.

\paragraph{Gradient tree boosting models.} The prediction of a \emph{single output
gradient boosting tree ensemble} is given by
\begin{equation}
f(x) = \rho_0 + \sum_{m=1}^M \mu \rho_m g_m(x),
\end{equation}
\noindent but also as
\begin{align}
f(x) = \sum_{m=1}^M \sum_{i=1}^n w^i_m(x) y^i = \sum_{i=1}^n w^i(x) y^i,
\end{align}
\noindent where the weight $w^i(x)$ takes into account the learning rate $\mu$,
the prediction of all tree models $g_m$ and the associated $\rho_m$. Given the
similarity between gradient boosting prediction and random forest model, we
deduce that the \emph{single target gradient boosting tree ensemble} has the
form of Equation~\ref{eq:br-dt-pred} and that \emph{multi-output gradient tree
boosting} (Algorithm~\ref{algo:gb-mo}) and \emph{gradient boosting tree with
projection of the output space and relabelling}
(Algorithm~\ref{algo:gbrt-rp-relabel}) has the form of
Equation~\ref{eq:mo-dt-pred}.

However, we note that the prediction model of the \emph{gradient tree boosting with
random projection of the output space} (Algorithm~\ref{algo:gbrt-rp}) is not
given by Equation~\ref{eq:br-dt-pred} and Equation~\ref{eq:mo-dt-pred} as the
prediction of a single output $j$ can combine the predictions of all $d$ outputs.
More formally, the prediction of the $j$-th output is given by:
\begin{equation}
f(x)_j = \sum_{m=1}^M \sum_{i=1}^n \sum_{k=1}^d w^i_{m,j,k}(x) y^i_k,
\end{equation}
\noindent where the weight function $w^i_{m,j,k}$ takes into account the contribution
of the $m$-th model fitted on a random projection $\phi_m$ of the output space
to predict the $j$-th output using the $k$-th outputs and the $i$-th sample. The triple
summation can be simplified by using a single weight to summarize the contribution of
all $M$ models:
\begin{align}
f(x)_j &= \sum_{m=1}^M \sum_{i=1}^n \sum_{k=1}^d w^i_{m,j,k}(x) y^i_k = \sum_{i=1}^n \sum_{k=1}^d w^i_{j,k}(x) y^i_k.
\end{align}

Between the studied methods, we can distinguish three groups of multi-output
tree models. The first one considers that all outputs are independent as with
binary relevance / single target trees, random forests or gradient
tree boosting models. The second group with multi-output random forests, gradient
boosting of multi-output tree and gradient boosting with random projection of
the output space and relabelling share the tree structures between all outputs,
but the leaf predictions are different for each output. The last and most
flexible group is the gradient tree boosting with random projection of the
output space sharing both the tree structures and the leaf predictions. We will
highlight in the experiments the impact of these differences in representation biases.

\section{Real world datasets}
\label{appendix:real-datasets}

Experiments are performed on several multi-label datasets:
the yeast~\cite{elisseeff2001kernel} and
the bird~\cite{briggs20139th}
datasets in the biology domain;
the corel5k~\cite{duygulu2002object} and
the scene~\cite{boutell2004learning}
datasets in the image domain;
the emotions~\cite{tsoumakas2008multi} and
the CAL500~\cite{turnbull2008semantic}
datasets in the music domain;
the bibtex~\cite{katakis2008multilabel},
the bookmarks~\cite{katakis2008multilabel},
the delicious~\cite{tsoumakas2008effective},
the enron~\cite{klimt2004enron},
the genbase~\cite{diplaris2005protein},
and the medical\footnote{The medical dataset comes from the computational medicine
    center's 2007 medical natural language processing challenge
    \url{http://computationalmedicine.org/challenge/previous}.}
datasets in the text domain
and the mediamill~\cite{snoek2006challenge}
dataset in the video domain.

Several hierarchical classification tasks are also included to increase
the diversity in the number of labels. They are treated as flat multi-label
classification tasks, with each node of the hierarchy treated as one
label. Nodes of the hierarchy which never occured in the training and
testing set were removed.  The reuters~\cite{rousu2005learning},
WIPO~\cite{rousu2005learning} datasets are from the text domain. The
Diatoms~\cite{dimitrovski2012hierarchical} dataset is from the image
domain.  SCOP-GO~\cite{clare2003machine},
Yeast-GO~\cite{barutcuoglu2006hierarchical} and
Expression-GO~\cite{vens2008decision} are from the biological domain.
Missing values in the Expression-GO dataset were inferred using the
median for continuous features and the most frequent value for
categorical features using the entire dataset.  The inference of a
drug-protein interaction network~\cite{yamanishi2011extracting} is
also considered either using the drugs to infer the interactions with
the protein (drug-interaction), either using the proteins to infer the
interactions with the drugs (protein-interaction).

Multi-output regression approaches are evaluated on several real world datasets:
the edm~\cite{karalivc1997first} dataset
in the industrial domain;
the water-quality \cite{dvzeroski2000predicting}
dataset in the environmental domain;
the atp1d~\cite{spyromitros2012multi},
the atp7d~\cite{spyromitros2012multi},
the scm1d~\cite{spyromitros2012multi} and
the scm20d~\cite{spyromitros2012multi}
datasets in the price prediction domain;
the oes97~\cite{spyromitros2012multi} and
the oes10~\cite{spyromitros2012multi} datasets
in the human resource domain. All outputs in these datasets were normalized
to have zero mean and unit variance.

\clearpage

\section{Performance of tree ensemble models over the multi-label datasets}
\label{appendix:overall-mulit-label}

Table~\ref{table:summary-multilabel} and Table~\ref{table:summary-multilabel-2}
show the performance of the random forest models and the boosting algorithms
over the 21 multi-label datasets.

% python analyse.py -g multilabel --to-latex
\begin{table}[h]
\caption{LRAP scores over 21 multi-label datasets (part 1).}
\label{table:summary-multilabel}
\centering
% \setlength{\tabcolsep}{1.5pt}
% \begin{small}\footnotesize
\begin{tabular}{llll}
\toprule
                            &                   CAL500 &                   bibtex &                  birds \\
\midrule
br-gbrt                     &  $0.505 \pm 0.002$ (3.5) &    $0.587 \pm 0.007$ (6) &  $0.787 \pm 0.009$ (6) \\
br-rf                       &    $0.484 \pm 0.002$ (8) &    $0.542 \pm 0.005$ (8) &  $0.802 \pm 0.013$ (1) \\
gbmort                      &    $0.501 \pm 0.005$ (6) &  $0.595 \pm 0.005$ (4.5) &  $0.772 \pm 0.007$ (8) \\
gbrt-relabel-rpo-gaussian   &    $0.507 \pm 0.009$ (1) &    $0.607 \pm 0.005$ (1) &    $0.800 \pm 0.017$ (2) \\
gbrt-relabel-rpo-subsampled &    $0.499 \pm 0.008$ (7) &    $0.596 \pm 0.005$ (3) &   $0.790 \pm 0.016$ (4) \\
gbrt-rpo-gaussian           &  $0.505 \pm 0.006$ (3.5) &      $0.600 \pm 0.003$ (2) &  $0.793 \pm 0.017$ (3) \\
gbrt-rpo-subsampled         &    $0.506 \pm 0.006$ (2) &  $0.595 \pm 0.007$ (4.5) &  $0.779 \pm 0.018$ (7) \\
mo-rf                       &    $0.502 \pm 0.003$ (5) &    $0.553 \pm 0.005$ (7) &  $0.789 \pm 0.012$ (5) \\
\midrule
                            &              bookmarks   &                corel5k &              delicious \\
\midrule
br-gbrt                     &  $0.4463 \pm 0.0038$ (7) &    $0.291 \pm 0.006$ (7) &    $0.347 \pm 0.002$ (8) \\
br-rf                       &  $0.4472 \pm 0.0019$ (6) &    $0.273 \pm 0.012$ (8) &  $0.373 \pm 0.004$ (6.5) \\
gbmort                      &  $0.4855 \pm 0.0016$ (2) &  $0.312 \pm 0.009$ (3.5) &  $0.384 \pm 0.003$ (3.5) \\
gbrt-relabel-rpo-gaussian   &  $0.4893 \pm 0.0003$ (1) &  $0.315 \pm 0.007$ (1.5) &    $0.389 \pm 0.003$ (1) \\
gbrt-relabel-rpo-subsampled &  $0.4718 \pm 0.0034$ (4) &     $0.310 \pm 0.007$ (5) &  $0.384 \pm 0.003$ (3.5) \\
gbrt-rpo-gaussian           &  $0.4753 \pm 0.0022$ (3) &   $0.315 \pm 0.010$ (1.5) &    $0.386 \pm 0.004$ (2) \\
gbrt-rpo-subsampled         &  $0.4621 \pm 0.0026$ (5) &  $0.312 \pm 0.006$ (3.5) &    $0.377 \pm 0.003$ (5) \\
mo-rf                       &  $0.4312 \pm 0.0023$ (8) &     $0.294 \pm 0.010$ (6) &  $0.373 \pm 0.004$ (6.5) \\
\midrule
                            &                diatoms &         drug-interaction &               emotions \\
\midrule
br-gbrt                     &  $0.623 \pm 0.007$ (7.5) &  $0.271 \pm 0.018$ (8) &      $0.800 \pm 0.022$ (7) \\
br-rf                       &  $0.623 \pm 0.011$ (7.5) &   $0.310 \pm 0.009$ (5) &    $0.816 \pm 0.009$ (1) \\
gbmort                      &    $0.656 \pm 0.012$ (4) &  $0.304 \pm 0.005$ (7) &    $0.794 \pm 0.014$ (8) \\
gbrt-relabel-rpo-gaussian   &     $0.725 \pm 0.010$ (1) &  $0.326 \pm 0.008$ (1) &  $0.802 \pm 0.017$ (5.5) \\
gbrt-relabel-rpo-subsampled &    $0.685 \pm 0.012$ (3) &  $0.322 \pm 0.009$ (3) &    $0.808 \pm 0.021$ (3) \\
gbrt-rpo-gaussian           &    $0.702 \pm 0.014$ (2) &  $0.323 \pm 0.011$ (2) &    $0.804 \pm 0.009$ (4) \\
gbrt-rpo-subsampled         &  $0.653 \pm 0.013$ (5.5) &  $0.312 \pm 0.013$ (4) &  $0.802 \pm 0.007$ (5.5) \\
mo-rf                       &   $0.653 \pm 0.010$ (5.5) &  $0.308 \pm 0.007$ (6) &      $0.810 \pm 0.010$ (2) \\
\midrule
                            &                    enron &                  genbase &                mediamill \\
\midrule
br-gbrt                     &    $0.685 \pm 0.006$ (6) &  $0.989 \pm 0.009$ (8) &   $0.7449 \pm 0.0020$ (8) \\
br-rf                       &    $0.683 \pm 0.005$ (7) &  $0.994 \pm 0.005$ (2) &  $0.7819 \pm 0.0009$ (1) \\
gbmort                      &  $0.705 \pm 0.004$ (2.5) &   $0.990 \pm 0.004$ (6) &  $0.7504 \pm 0.0013$ (7) \\
gbrt-relabel-rpo-gaussian   &  $0.705 \pm 0.003$ (2.5) &  $0.993 \pm 0.006$ (3) &   $0.7660 \pm 0.0021$ (3) \\
gbrt-relabel-rpo-subsampled &    $0.697 \pm 0.004$ (5) &    $0.990 \pm 0.010$ (6) &  $0.7588 \pm 0.0013$ (5) \\
gbrt-rpo-gaussian           &    $0.706 \pm 0.004$ (1) &  $0.992 \pm 0.007$ (4) &  $0.7608 \pm 0.0008$ (4) \\
gbrt-rpo-subsampled         &    $0.699 \pm 0.005$ (4) &   $0.990 \pm 0.005$ (6) &  $0.7519 \pm 0.0006$ (6) \\
mo-rf                       &    $0.676 \pm 0.004$ (8) &  $0.995 \pm 0.004$ (1) &  $0.7793 \pm 0.0015$ (2) \\
\bottomrule
\end{tabular}
% \end{small}
\end{table}

\clearpage

\begin{table}[!h]
\caption{LRAP scores over 21 multi-label datasets (part 2).}
\label{table:summary-multilabel-2}
\centering
% \setlength{\tabcolsep}{1.5pt}
% \begin{small}\footnotesize
\begin{tabular}{llll}
\toprule
                            &                medical &      protein-interaction &                  reuters \\
\midrule
br-gbrt                     &    $0.864 \pm 0.006$ (3) &    $0.294 \pm 0.007$ (6) &   $0.939 \pm 0.0033$ (7) \\
br-rf                       &    $0.821 \pm 0.007$ (8) &    $0.293 \pm 0.006$ (7) &  $0.9406 \pm 0.0016$ (6) \\
gbmort                      &  $0.867 \pm 0.011$ (1.5) &   $0.310 \pm 0.007$ (2.5) &  $0.9483 \pm 0.0014$ (3) \\
gbrt-relabel-rpo-gaussian   &  $0.867 \pm 0.019$ (1.5) &   $0.310 \pm 0.009$ (2.5) &  $0.9508 \pm 0.0009$ (1) \\
gbrt-relabel-rpo-subsampled &    $0.856 \pm 0.012$ (5) &  $0.303 \pm 0.003$ (4.5) &  $0.9441 \pm 0.0016$ (4) \\
gbrt-rpo-gaussian           &    $0.859 \pm 0.017$ (4) &    $0.311 \pm 0.007$ (1) &  $0.9486 \pm 0.0021$ (2) \\
gbrt-rpo-subsampled         &    $0.851 \pm 0.009$ (6) &  $0.303 \pm 0.003$ (4.5) &   $0.9430 \pm 0.0031$ (5) \\
mo-rf                       &    $0.827 \pm 0.006$ (7) &    $0.288 \pm 0.009$ (8) &  $0.9337 \pm 0.0021$ (8) \\
\midrule
                            &                    scene &                  scop-go &        sequence-funcat \\
\midrule
br-gbrt                     &     $0.880 \pm 0.003$ (4) &  $0.716 \pm 0.047$ (8) &  $0.678 \pm 0.008$ (6) \\
br-rf                       &    $0.876 \pm 0.003$ (6) &  $0.798 \pm 0.004$ (2) &  $0.658 \pm 0.008$ (7) \\
gbmort                      &    $0.886 \pm 0.004$ (1) &  $0.796 \pm 0.007$ (3) &  $0.699 \pm 0.005$ (3) \\
gbrt-relabel-rpo-gaussian   &  $0.884 \pm 0.006$ (2.5) &  $0.788 \pm 0.006$ (4) &  $0.703 \pm 0.007$ (2) \\
gbrt-relabel-rpo-subsampled &    $0.879 \pm 0.008$ (5) &    $0.770 \pm 0.010$ (6) &  $0.685 \pm 0.008$ (5) \\
gbrt-rpo-gaussian           &  $0.884 \pm 0.005$ (2.5) &  $0.775 \pm 0.018$ (5) &  $0.706 \pm 0.007$ (1) \\
gbrt-rpo-subsampled         &    $0.875 \pm 0.006$ (7) &  $0.723 \pm 0.016$ (7) &  $0.691 \pm 0.006$ (4) \\
mo-rf                       &    $0.865 \pm 0.003$ (8) &    $0.800 \pm 0.006$ (1) &  $0.643 \pm 0.003$ (8) \\
\midrule
                            &                   wipo &                    yeast &                 yeast-go \\
\midrule
br-gbrt                     &  $0.706 \pm 0.009$ (6) &    $0.756 \pm 0.009$ (8) &  $0.499 \pm 0.009$ (4.5) \\
br-rf                       &  $0.633 \pm 0.013$ (7) &   $0.760 \pm 0.008$ (3.5) &     $0.463 \pm 0.010$ (7) \\
gbmort                      &  $0.762 \pm 0.011$ (3) &   $0.760 \pm 0.007$ (3.5) &    $0.504 \pm 0.015$ (3) \\
gbrt-relabel-rpo-gaussian   &  $0.776 \pm 0.012$ (1) &    $0.762 \pm 0.007$ (2) &    $0.524 \pm 0.012$ (1) \\
gbrt-relabel-rpo-subsampled &  $0.751 \pm 0.017$ (4) &  $0.758 \pm 0.005$ (5.5) &    $0.496 \pm 0.013$ (6) \\
gbrt-rpo-gaussian           &   $0.763 \pm 0.010$ (2) &    $0.763 \pm 0.005$ (1) &    $0.522 \pm 0.012$ (2) \\
gbrt-rpo-subsampled         &  $0.724 \pm 0.011$ (5) &  $0.758 \pm 0.008$ (5.5) &  $0.499 \pm 0.011$ (4.5) \\
mo-rf                       &  $0.624 \pm 0.018$ (8) &    $0.757 \pm 0.008$ (7) &    $0.415 \pm 0.014$ (8) \\
\bottomrule
\end{tabular}
% \end{small}
\end{table}

\clearpage

\section{Performance of tree ensemble models over the multi-output regression datasets}
\label{appendix:overall-multi-output-regression}

Table~\ref{table:summary-regression} shows the performance of the random forest
models and the boosting algorithms over the 8 multi-output regression datasets.

% python analyse.py -g regression --to-latex

\begin{table}[h]
\caption{Performance over 8 multi-output regression dataset}
\label{table:summary-regression}
\centering
% \setlength{\tabcolsep}{1.5pt}
% \begin{small}\footnotesize
\begin{tabular}{llll}
\toprule
                            &                  atp1d &                  atp7d &                    edm \\
\midrule
gbmort                      &  $0.80 \pm 0.03$ (5.5) &  $0.63 \pm 0.03$ (2) &  $0.39 \pm 0.16$ (3) \\
gbrt-relabel-rpo-gaussian   &  $0.81 \pm 0.03$ (3.5) &  $0.66 \pm 0.04$ (1) &  $0.25 \pm 0.28$ (8) \\
gbrt-relabel-rpo-subsampled &  $0.79 \pm 0.04$ (7)   &  $0.54 \pm 0.13$ (7) &  $0.35 \pm 0.10$ (5) \\
gbrt-rpo-gaussian           &  $0.80 \pm 0.04$ (5.5) &  $0.54 \pm 0.20$ (7) &  $0.36 \pm 0.04$ (4) \\
gbrt-rpo-subsampled         &  $0.81 \pm 0.04$ (3.5) &  $0.54 \pm 0.16$ (7) &  $0.31 \pm 0.27$ (7) \\
mo-rf                       &  $0.82 \pm 0.03$ (2)   &  $0.60 \pm 0.06$ (4) &  $0.51 \pm 0.02$ (1) \\
st-gbrt                     &  $0.78 \pm 0.05$ (8)   &  $0.59 \pm 0.08$ (5) &  $0.34 \pm 0.14$ (6) \\
st-rf                       &  $0.83 \pm 0.02$ (1)   &  $0.61 \pm 0.07$ (3) &  $0.47 \pm 0.04$ (2) \\
\midrule
                            &                  oes10 &                  oes97 &                    scm1d \\
\midrule
gbmort                      &  $0.77 \pm 0.05$ (3.5) &    $0.67 \pm 0.07$ (8)   & $0.908 \pm 0.003$ (4.5) \\
gbrt-relabel-rpo-gaussian   &  $0.75 \pm 0.04$ (7.5) &    $0.71 \pm 0.07$ (2.5) & $0.910 \pm 0.004$ (2.5) \\
gbrt-relabel-rpo-subsampled &  $0.75 \pm 0.06$ (7.5) &    $0.68 \pm 0.07$ (6)   & $0.912 \pm 0.003$ (1) \\
gbrt-rpo-gaussian           &  $0.77 \pm 0.03$ (3.5) &    $0.68 \pm 0.08$ (6)   & $0.910 \pm 0.004$ (2.5) \\
gbrt-rpo-subsampled         &  $0.76 \pm 0.02$ (5.5) &    $0.71 \pm 0.08$ (2.5) & $0.908 \pm 0.004$ (4.5) \\
mo-rf                       &  $0.76 \pm 0.04$ (5.5) &    $0.69 \pm 0.05$ (4)   & $0.898 \pm 0.004$ (8) \\
st-gbrt                     &  $0.79 \pm 0.03$ (1.5) &    $0.68 \pm 0.07$ (6)   & $0.905 \pm 0.003$ (7) \\
st-rf                       &  $0.79 \pm 0.03$ (1.5) &    $0.72 \pm 0.05$ (1)   & $0.907 \pm 0.004$ (6) \\
\midrule
                            &                   scm20d &          water-quality \\
\midrule
gbmort                      &  $0.856 \pm 0.006$ (2) &    $0.14 \pm 0.01$ (4.5) \\
gbrt-relabel-rpo-gaussian   &  $0.862 \pm 0.006$ (1) &    $0.15 \pm 0.01$ (2) \\
gbrt-relabel-rpo-subsampled &  $0.854 \pm 0.007$ (3) &    $0.14 \pm 0.02$ (4.5) \\
gbrt-rpo-gaussian           &  $0.852 \pm 0.006$ (4) &    $0.14 \pm 0.01$ (4.5) \\
gbrt-rpo-subsampled         &  $0.850 \pm 0.007$ (5) &    $0.13 \pm 0.02$ (7.5) \\
mo-rf                       &  $0.849 \pm 0.007$ (6.5) &  $0.16 \pm 0.01$ (1) \\
st-gbrt                     &  $0.836 \pm 0.006$ (8)   &  $0.13 \pm 0.02$ (7.5) \\
st-rf                       &  $0.849 \pm 0.006$ (6.5) &  $0.14 \pm 0.01$ (4.5) \\
\bottomrule
\end{tabular}
% \end{small}
\end{table}

\bibliographystyle{spbasic}      % basic style, author-year citations
\bibliography{bibliography}   % name your BibTeX data base

\end{document}